%% file: bopp.tex
\documentclass[]{article}
\PassOptionsToPackage{numbers,compress}{natbib}
\usepackage{jmlr2e}
\usepackage{url}
\usepackage{amsmath}
\usepackage{amssymb}
\usepackage{amsbsy}
\usepackage{siunitx}
\usepackage{algorithm}
\usepackage{algorithmicx}
\usepackage{algpseudocode}
\usepackage{mathtools}
\usepackage{array}
\usepackage{setspace}
\usepackage{subcaption}
\usepackage{booktabs}
\usepackage{adjustbox}
\usepackage{microtype}

\usepackage{times}
\usepackage{footmisc}
\usepackage{listings}
\usepackage{color}
\usepackage{xcolor}
\usepackage{textcomp}
\usepackage{xspace}

\usepackage{fancyhdr}

\usepackage{bibentry}
\nobibliography*

\usepackage{paralist}

\usepackage{natbib}

\usepackage{enumitem}
\setlist[itemize]{leftmargin=*}

\usepackage{hyperref}

\definecolor{darkgreen}{rgb}{0.25,.5,0}
\definecolor{blue}{rgb}{0,0.33,0.66}
\definecolor{red}{rgb}{0.66,0.0,0.0}
\definecolor{purple}{rgb}{0.33,0,0.66}
\definecolor{cyan}{rgb}{0.0,0.5,0.5}
\definecolor{orange}{rgb}{0.5,0.25,0.0}
\definecolor{gray}{rgb}{0.4,0.4,0.4}
\lstnewenvironment{code}[2]{\lstset{caption=#1,label=#2}}{}

\newcommand{\lsi}[1]{\lstinline$#1$}

\newcommand{\sample}{\lstinline$sample$\xspace}
\newcommand{\observe}{\lstinline$observe$\xspace}
\newcommand{\observes}{\lstinline$observe<-$\xspace}

\newcommand{\simulatec}{\lstinline$simulate$\xspace}
\newcommand{\abcl}{\lstinline$abc-likelihood$\xspace}
\newcommand{\factor}{\lstinline$factor$\xspace}

\newcommand{\defquery}{\lstinline$defquery$\xspace}
\newcommand{\query}{\lstinline$query$\xspace}
\newcommand{\defopt}{\lstinline$defopt$\xspace}
\newcommand{\doquery}{\lstinline$doquery$\xspace}
\newcommand{\doopt}{\lstinline$doopt$\xspace}

\newcolumntype{A}{>{\centering\arraybackslash}m{1cm}}

\newcolumntype{B}{>{\centering\arraybackslash}m{1.5cm}}

\newcolumntype{D}{>{\centering\arraybackslash}m{2cm}}

\algnewcommand{\IIf}[1]{\unskip \algorithmicif\ #1\ \algorithmicthen}
\algnewcommand{\ElseIIf}{\unskip\ \algorithmicelse}
\algnewcommand{\EndIIf}{\unskip\ \algorithmicend\ \algorithmicif}

\newcommand{\real}{\mathbb{R}}

\newcommand{\hth}{\hat{\theta}}

\newcommand{\RN}[1]{%
	\textup{\uppercase\expandafter{\romannumeral#1}}%
}
\renewcommand{\v}[1]{\ensuremath{\boldsymbol{#1}}}

\DeclareMathOperator*{\argmax}{arg\,\!max}

\DeclareMathOperator{\Dirichlet}{Dirichlet}

\newcommand{\qmarg}{\texttt{q-marg}}
\newcommand{\qprior}{\texttt{q-prior}}
\newcommand{\qacq}{\texttt{q-acq}}

\ShortHeadings{Bayesian Optimization for Probabilistic Programs}{Rainforth, Le, van de Meent, Osborne and Wood}
\firstpageno{1}

\title{Bayesian Optimization for Probabilistic Programs$^*$}

\author{\name Tom Rainforth \email twgr@robots.ox.ac.uk \\
	\addr Department of Engineering Science, University of Oxford
	\AND
	\name Tuan Anh Le \email tuananh@robots.ox.ac.uk \\
	\addr Department of Engineering Science, University of Oxford
	\AND
	\name Jan-Willem van de Meent \email j.vandemeent@northeastern.edu \\
	\addr College of Computer and Information Science, Northeastern University
	\AND
	\name Michael A Osborne \email mosb@robots.ox.ac.uk \\
	\addr Department of Engineering Science, University of Oxford
	\AND
	\name Frank Wood \email fwood@robots.ox.ac.uk \\
	\addr Department of Engineering Science, University of Oxford \\
}

\fancypagestyle{empty}{%
	\fancyhf{}
	\fancyhead[L]{}
	\renewcommand{\headrulewidth}{0pt}
	\renewcommand{\footrulewidth}{0pt}
	\fancyfoot[L]{\footnotesize \vspace{-25pt} \rule{\textwidth}{0.4pt} \indent{} \indent{} *Please cite this version: \\ \vspace{5pt} \hspace{20pt} \begin{minipage}{0.9\textwidth}\bibentry{rainforth2016bayesian}\end{minipage}}
}

\begin{document}
%

\maketitle
\vspace{-30pt}

\newcommand\blfootnote[1]{%
	\begingroup
	\renewcommand\thefootnote{}\footnote{#1}%
	\addtocounter{footnote}{-1}%
	\endgroup
}

\begin{abstract}
\input{abstract}
\end{abstract}

\thispagestyle{empty}

\section{Introduction} 
\label{sec:IntroductionBOPP}

\input{introduction}

\section{Background}

\input{probprog}

\input{bayesOpt}

\section{Problem Formulation}
\label{sec:problem}

\input{problem-formulation.tex}

\section{Bayesian Program Optimization}
\label{sec:bopp}

\input{optimization}

\section{Experiments}

\input{experiments.tex}

\section{Discussion and Future Work}
\label{sec:disc}

\input{discussion}

\newpage

\appendix

\section{Program Transformations in Detail}
\label{sec:program-transformations}
\input{program-transformations}

\section{Problem Independent Gaussian Process Hyperprior}
\label{sec:app:hyperprior}

\input{gp.tex}

\section{Full Details for House Heating Experiment}
\label{sec:app:heating}

\input{house-heating.tex}

\newpage

\section*{Acknowledgements}
Tom Rainforth is supported by a BP industrial grant. Tuan Anh Le is supported by a Google studentship, project code DF6700.  Frank Wood is supported under DARPA PPAML through the U.S. AFRL under Cooperative Agreement FA8750-14-2-0006, Sub Award number 61160290-111668.

\bibliography{refsBOPP}{}

\end{document}

%% file: abstract.tex

We present the first general purpose framework for marginal maximum a posteriori estimation of probabilistic program variables. By using a series of code transformations, the evidence of any probabilistic program, and therefore of any graphical model, can be optimized with respect to an arbitrary subset of its sampled variables.  To carry out this optimization, we develop the first Bayesian optimization package to directly exploit the source code of its target, leading to innovations in problem-independent hyperpriors, unbounded optimization, and implicit constraint satisfaction; delivering significant performance improvements over prominent existing packages.  We present applications of our method to a number of tasks including engineering design and parameter optimization.

%% file: introduction.tex

Probabilistic programming systems (PPS) allow probabilistic models to be represented in the form of a generative model and statements for conditioning on data \citep{carpenter2015stan,goodman_uai_2008,goodman_book_2014,mansinghka2014venture,minka_software_2010,wood2014new}.  
Their core philosophy is to decouple model specification and inference, the former corresponding to the user-specified program code and the latter to an inference engine capable of operating on arbitrary programs.  Removing the need for users to write inference algorithms significantly reduces the burden of developing new models and makes effective statistical methods accessible to non-experts.

Although significant progress has been made on the problem of general purpose \emph{inference} of program variables, less attention has been given to their \emph{optimization}.  Optimization is an essential tool for effective machine learning, necessary when the user requires a single estimate. It also often forms a tractable alternative when full inference is infeasible \citep{murphy2012machine}.  Moreover, coincident optimization and inference is often required, corresponding to a marginal maximum a posteriori (MMAP) setting where one wishes to maximize some variables, while marginalizing out others.  Examples of MMAP problems include hyperparameter optimization, expectation maximization, and policy search \citep{van2015black}.

In this paper we develop the first system that extends probabilistic programming (PP) to this more general MMAP framework, wherein the user specifies a model in the same manner as existing systems, but then selects some subset of the sampled variables in the program to be optimized, with the rest marginalized out using existing inference algorithms.  The \textit{optimization query} we introduce can be implemented and utilized in any PPS that supports an inference method returning a marginal likelihood estimate.  This framework increases the scope of models that can be expressed in PPS and gives additional flexibility in the outputs a user can request from the program.

MMAP estimation is difficult as it corresponds to the optimization of an intractable integral, such that the optimization target is expensive to evaluate and gives noisy results.  Current PPS inference engines are typically unsuited to such settings.  We therefore introduce BOPP\footnote{Code available at \href{http://www.github.com/probprog/bopp/}{\url{http://www.github.com/probprog/bopp/}} \vspace{7pt}} 
(Bayesian optimization for probabilistic programs) which couples existing inference algorithms from PPS, like \emph{Anglican} \citep{wood2014new}, with a new Gaussian process (GP) \citep{rasmussen2006gaussian} based Bayesian optimization (BO) package \citep{gutmann2016bayesian, jones1998efficient, osborne2009gaussian, shahriari2016unbounded}.  

To demonstrate the functionality provided by BOPP, we consider an example application of engineering design.  Engineering design relies extensively on simulations which typically have two things in common: the desire of the user to find a single best design and an uncertainty in the environment in which the designed component will live. Even when these simulations are deterministic, this is an approximation to a truly stochastic world. By expressing the utility of a particular design-environment combination using an approximate Bayesian computation (ABC) likelihood \citep{csillery2010approximate}, one can pose this as a MMAP problem, optimizing the design while marginalizing out the environmental uncertainty.

\flushbottom

\begin{figure*}[p]
	\centering
	\begin{subfigure}[t]{0.47\textwidth}
		\includegraphics[width=\textwidth]{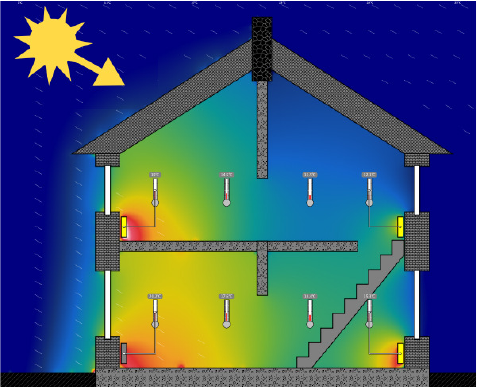}
		\caption{Radiator powers set evenly}
	\end{subfigure}
	~~~~ 
		\begin{subfigure}[t]{0.47\textwidth}
			\includegraphics[width=\textwidth]{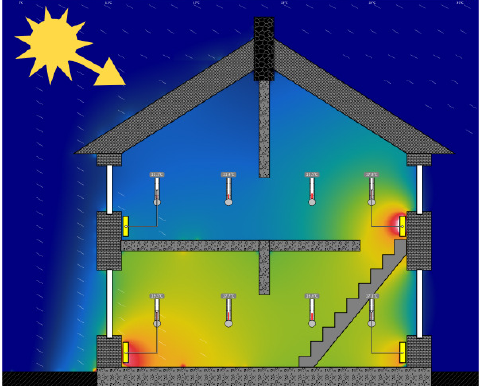}
			\caption{Best setup from BOPP initialization}
		\end{subfigure} \\
		\vspace{10pt}
			\begin{subfigure}[t]{0.47\textwidth}
				\includegraphics[width=\textwidth]{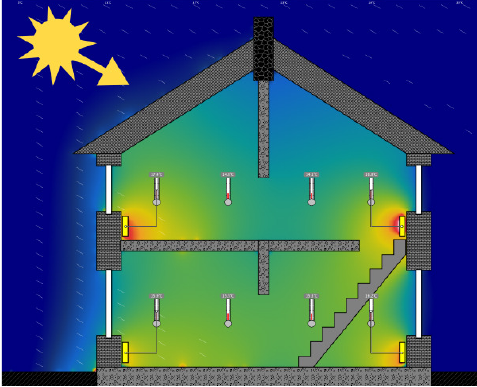}
				\caption{Best setup after 100 iterations of BOPP}
			\end{subfigure}
		~~~~ 
			\begin{subfigure}[t]{0.47\textwidth}
				\includegraphics[width=\textwidth]{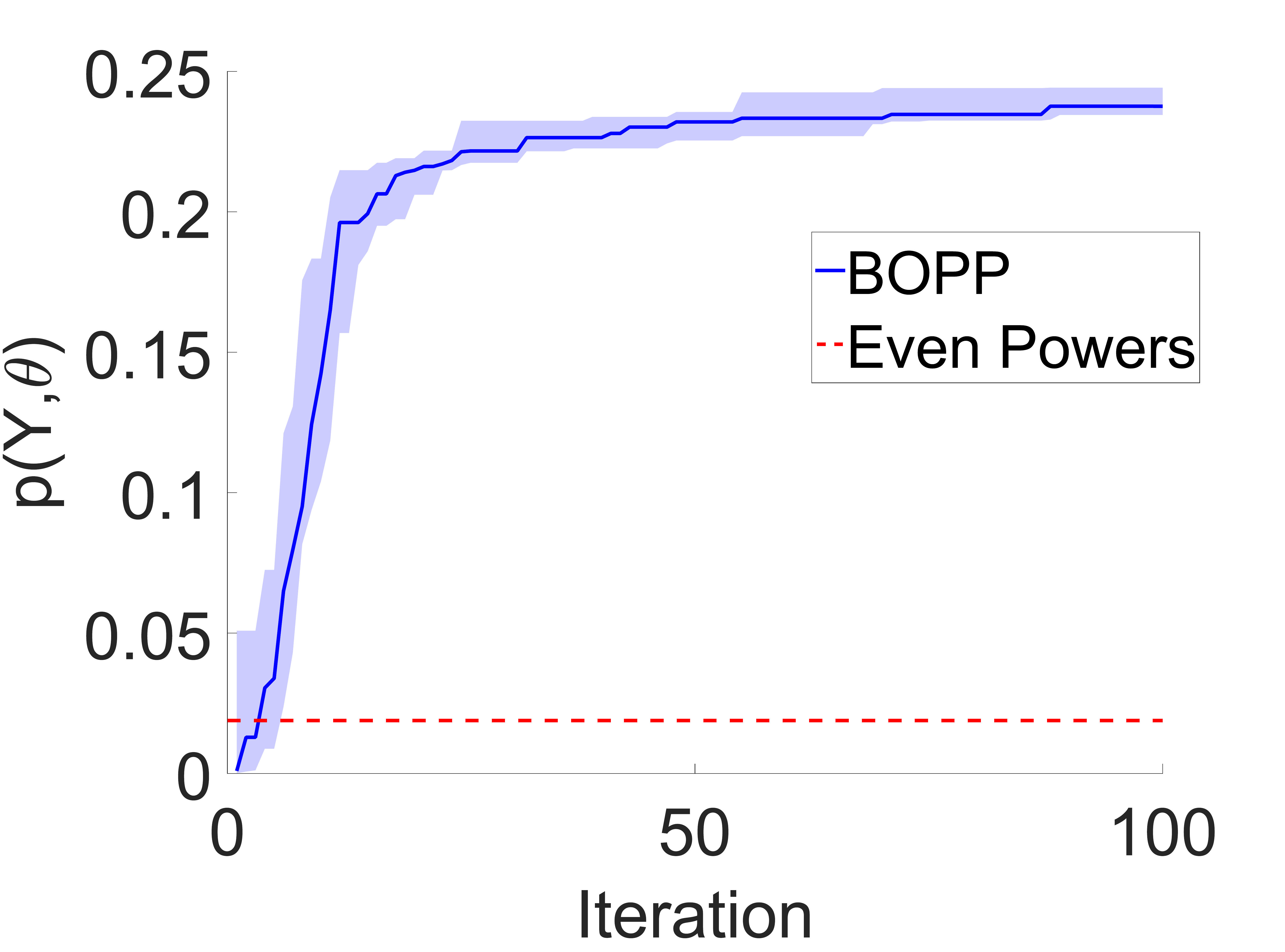}
				\caption{Convergence of evidence}
			\end{subfigure}
	%
	\caption{
		\label{fig:houses}
		Simulation-based optimization of radiator powers subject to varying solar intensity. Shown are output heat maps from Energy2D \citep{xie2012energy2d} simulations at one intensity, corresponding to setting all the radiators to the same power (\emph{top left}), the best result from a set of 5 randomly chosen powers used for initializing BOPP (\emph{top right}), and the best setup found after 100 iterations of BOPP (\emph{bottom left}). The bottom right plot shows convergence of the evidence of the respective model, giving the median and 25/75\% quartiles.
	}
\end{figure*}

\begin{figure}[p]
	\begin{lstlisting}[basicstyle=\ttfamily]
(defopt house-heating [alphas target-temperatures] [powers]
 (let [solar-intensity (sample weather-prior)
       powers (sample (dirichlet alphas))
       temperatures (simulate solar-intensity powers)]
  (observe (abc-likelihood temperatures) target-temperatures)))
	\end{lstlisting}	
	\vspace{-6pt}
	\caption{BOPP query for optimizing the power allocation to radiators in a house.  Here \lstinline{weather-prior} is a distribution over the solar intensity and a uniform Dirichlet prior with concentration \lsi{alpha} is placed over the powers. Calling \simulatec performs an Energy2D simulation of house temperatures. The utility of the resulting output is incorporated using \abcl, which measures a discrepency from the \texttt{target-temperatures}. Calling \doopt on this query invokes the BOPP algorithm to perform MMAP estimation, where the second input \lstinline{powers} indicates the variable to be optimized. \label{fig:house-heating-code}}
\end{figure}

Figure~\ref{fig:houses} illustrates how BOPP can be applied to engineering design, taking the example of optimizing the distribution of power between radiators in a house so as to homogenize the temperature, while marginalizing out possible weather conditions and subject to a total energy budget. The probabilistic program shown in Figure~\ref{fig:house-heating-code} allows us to define a prior over the uncertain weather, while conditioning on the output of a deterministic simulator (here Energy2D \citep{xie2012energy2d}-a finite element package for heat transfer) using an ABC likelihood.  BOPP now allows the required coincident inference and optimization to be carried out automatically, directly returning increasingly optimal configurations. 

BO is an attractive choice for the required optimization in MMAP as it is typically efficient in the number of target evaluations, operates on non-differentiable targets, and incorporates noise in the target function evaluations.  However, applying BO to probabilistic programs presents challenges, such as the need to give robust performance on a wide range of problems with varying scaling and potentially unbounded support.  Furthermore, the target program may contain unknown constraints, implicitly defined by the generative model, and variables whose type is unknown (i.e. they may be continuous or discrete).

On the other hand, the availability of the target source code in a PPS presents opportunities to overcome these issues and go beyond what can be done with existing BO packages.  BOPP exploits the source code in a number of ways, such as optimizing the acquisition function using the original generative model to ensure the solution satisfies the implicit constaints, performing adaptive domain scaling to ensure that GP kernel hyperparameters can be set according to problem-independent hyperpriors, and defining an adaptive non-stationary mean function to support unbounded BO. 

Together, these innovations mean that BOPP can be run in a manner that is fully black-box from the user's perspective, requiring only the identification of the target variables relative to current syntax for operating on arbitrary programs. We further show that BOPP is competitive with existing BO engines for direct optimization on common benchmarks problems that do not require marginalization.

%% file: probprog.tex

\subsection{Probabilistic Programming}
\label{sec:prob-prog}

Probabilistic programming systems allow users to define probabilistic models using a domain-specific programming language. A probabilistic program implicitly defines a distribution on random variables, whilst the system back-end implements general-purpose inference methods.  

PPS such as Infer.Net \citep{minka_software_2010} and Stan \citep{carpenter2015stan} can be thought of as defining graphical models or factor graphs.  Our focus will instead be on systems such as Church \citep{goodman_uai_2008}, Venture \citep{mansinghka2014venture}, WebPPL \citep{goodman_book_2014}, and Anglican \citep{wood2014new}, which employ a general-purpose programming language for model specification. In these systems, the set of random variables is dynamically typed, such that it is possible to write programs in which this set differs from execution to execution.  This allows an unspecified number of random variables and incorporation of arbitrary black box deterministic functions, such as was exploited by the \simulatec function in Figure \ref{fig:house-heating-code}. The price for this expressivity is that inference methods must be formulated in such a manner that they are applicable to models where the density function is intractable and can only be evaluated during forwards simulation of the program. 

One such general purpose system, \emph{Anglican}, will be used as a reference in this paper.  In Anglican, models are defined using the inference macro \defquery. These models, which we refer to as queries \citep{goodman_uai_2008}, specify a joint distribution $p(Y,X)$ over data $Y$ and variables $X$. Inference on the model is performed using the macro \doquery, which produces a sequence of approximate samples from the conditional distribution $p(X|Y)$ and, for importance sampling based inference algorithms (e.g. sequential Monte Carlo), a marginal likelihood estimate $p(Y)$.  

Random variables in an Anglican program are specified using \sample statements, which can be thought of as terms in the prior. Conditioning is specified using \observe statements which can be thought of as likelihood terms.  Outputs of the program, taking the form of posterior samples, are indicated by the return values.  There is a finite set of \sample and \observe statements in a program source code, but the number of times each statement is called can vary between executions.  We refer the reader to  \href{http://www.robots.ox.ac.uk/~fwood/anglican/}{\small\url{http://www.robots.ox.ac.uk/~fwood/anglican/}} for more details.

%% file: bayesOpt.tex
\vspace{10pt}
\subsection{Bayesian Optimization}
\label{sec:BO}
Consider an arbitrary black-box target function $f \colon \vartheta \rightarrow \real$ that can be evaluated for an arbitrary point $\theta \in \vartheta$ to produce, potentially noisy, outputs $\hat{w} \in \real$.  BO \citep{jones1998efficient,osborne2009gaussian} aims to find the global maximum
\begin{align}
\label{eq:funcMax}
\theta^* = \argmax_{\theta \in \vartheta} f\left(\theta\right).
\end{align}
The key idea of BO is to place a prior on $f$ that expresses belief about the space of functions within which $f$ might live.  When the function is evaluated, the resultant information is incorporated by conditioning upon the observed data to give a posterior over functions.  
This allows estimation of the expected value and uncertainty in $f\left(\theta\right)$ for all $\theta \in \vartheta$.  
From this, an acquisition function $\zeta : \vartheta \rightarrow \real$ is defined, which assigns an expected utility to evaluating $f$ at particular $\theta$, based on the trade-off between exploration and exploitation in finding the maximum.  When direct evaluation of $f$ is expensive, the acquisition function constitutes a cheaper to evaluate substitute, which is optimized to ascertain the next point at which the target function should be evaluated in a sequential fashion.  By interleaving optimization of the acquisition function, evaluating $f$ at the suggested point, and updating the surrogate, BO forms a global optimization algorithm that is typically very efficient in the required number of function evaluations, whilst naturally dealing with noise in the outputs.  Although alternatives such as random forests \citep{bergstra2011algorithms,hutter2011sequential} or neural networks \citep{snoek2015scalable} exist, the most common prior used for $f$ is a GP \citep{rasmussen2006gaussian}.  
For further information on BO we refer the reader to the recent review by Shahriari et al \cite{shahriari2016taking}.

\subsection{Gaussian Processes}
\label{sec:GPs}

Informally one can think of a Gaussian Process (GP) \citep{rasmussen2006gaussian} as being a nonparametric distribution over functions which is fully specified by a mean function $\mu \colon \vartheta \rightarrow \real$ and covariance function $k \colon \vartheta \times \vartheta \rightarrow \real$, the latter of which must be a bounded $\left(\text{i.e. }k\left(\theta,\theta'\right)<\infty, \; \forall \theta,\theta' \in \vartheta\right)$ and reproducing kernel.  We can describe a function $f$ as being distributed according to a GP:
\begin{align}
\label{eq:GP}
f \left(\theta\right) \sim GP \left(\mu\left(\theta\right), k\left(\theta,\theta'\right)\right)
\end{align}
which by definition means that the functional evaluations realized at any finite number of sample points is distributed according to a multivariate Gaussian. Note that the inputs to $\mu$ and $k$ need not be numeric and as such a GP can be defined over anything for which kernel can be defined.

An important property of a GP is that it is conjugate with a Gaussian likelihood.  Consider pairs of input-output data points $\{\hth_j,\hat{w}_j\}_{j=1:m}$, $\hat{W} = \{\hat{w}_j\}_{j=1:m}$, $\hat{\Theta} = \{\hth_j\}_{j=1:m}$ and the separable likelihood function
\begin{align}
\label{eq:GP-lik}
p(\hat{W}| \hat{\Theta}, f) = \prod_{j=1}^{m}p(\hat{w}_j | f(\hth_j)) = \prod_{j=1}^{m}\frac{1}{\sigma_{n}\sqrt{2\pi}} \exp \left(-\frac{\left(\hat{w}_j-f(\hth_j)\right)^2}{2\sigma_n^2}\right)
\end{align}
where $\sigma_n$ is an observation noise. Using a GP prior $f\left(\theta\right)\sim GP(\mu_{\text{prior}}\left(\theta\right),k_{\text{prior}}\left(\theta,\theta\right))$ leads to an analytic GP posterior 
\begin{align}
\label{eq:gpPosterior}
\mu_{post} \left(\theta\right) & = \mu_{\text{prior}}\left(\theta\right) + k_{\text{prior}}\left(\theta,\hat{\Theta} \right) \left[k_{\text{prior}}\left(\hat{\Theta} ,\hat{\Theta}  \right) + \sigma_n^2 I\right]^{-1} \left(\hat{W} -\mu_{\text{prior}}\left(\hat{\Theta} \right)\right) \\
k_{post} \left(\theta,\theta'\right) & = k_{\text{prior}} \left(\theta,\theta'\right) - k_{\text{prior}}\left(\theta,\hat{\Theta} \right) \left[k_{\text{prior}}\left(\hat{\Theta},\hat{\Theta} \right) + \sigma_n^2 I\right]^{-1} k_{\text{prior}}\left(\hat{\Theta} ,\theta'\right)
\end{align}
and Gaussian predictive distribution
\begin{align}
\label{eq:gpPred}
w | \theta, \hat{W}, \hat{\Theta} \sim \mathcal{N} \left(\mu_{post}\left(\theta\right), k_{post} \left(\theta,\theta\right) + \sigma_n^2 I\right)
\end{align}
where we have used the shorthand $k_{\text{prior}}(\hat{\Theta},\hat{\Theta}) = \left[\begin{smallmatrix} k_{\text{prior}}(\hth_1,\hth_1) & k_{\text{prior}}(\hth_1,\hth_2) & \dots\\ k_{\text{prior}}(\hth_2,\hth_1) & k_{\text{prior}}(\hth_2,\hth_2) & \dots \\ \dots & \dots & \dots\end{smallmatrix}\right]$ and similarly for $\mu_{\text{prior}}$, $\mu_{\text{post}}$ and $k_{\text{post}}$.

%% file: problem-formulation.tex

Given a program defining the joint density $p(Y, X, \theta)$ with fixed $Y$, our aim is to optimize with respect to a subset of the variables $\theta$ whilst marginalizing out latent variables $X$
\begin{equation}
\label{eq:MMAP}
\theta^* = \argmax_{\theta \in \vartheta} \; p(\theta | Y) = \argmax_{\theta \in \vartheta} \; p(Y, \theta) = \argmax_{\theta \in \vartheta} \: \int \: p(Y, X, \theta) dX.
\end{equation}

To provide syntax to differentiate between $\theta$ and $X$, we introduce a new query macro \defopt.  The syntax of \defopt is identical to \defquery except that it has an additional input identifying the variables to be optimized.  To allow for the interleaving of inference and optimization required in MMAP estimation, we further introduce \doopt, which, analogous to \doquery, returns a lazy sequence $\{\hat{\theta}^*_m,\hat{\Omega}^*_m,\hat{u}^*_m\}_{m=1,\dots}$ where $\hat{\Omega}^*_m \subseteq X$ are the program outputs associated with $\theta=\hth^*_m$ and each $\hat{u}^*_m \in \real^+$ is an estimate of the corresponding log marginal $\log p(Y, \hth_m^*)$ (see Section \ref{sec:bopp-for-ml}).  The sequence is defined such that, at any time, $\hat{\theta}^*_m$ corresponds to the point expected to be most optimal of those evaluated so far and allows both inference and optimization to be carried out online.

Although no restrictions are placed on $X$, it is necessary to place some restrictions on how programs  use the optimization variables $\theta = \phi_{1:K}$ specified by the optimization argument list of \defopt.
First, each optimization variable $\phi_k$ must be bound to a value directly by a \sample statement with fixed measure-type distribution argument.
This avoids change of variable complications arising from nonlinear deterministic mappings.  
Second, in order for the optimization to be well defined, the program must be written such that any possible execution trace binds each optimization variable $\phi_k$ exactly once.  
Finally, although any $\phi_k$ may be lexically multiply bound, it must have the same base measure in all possible execution traces, because, for instance, if the base measure of a $\phi_k$ were to change from Lebesgue to counting, the notion of optimality would no longer admit a conventional interpretation.
Note that although the transformation implementations shown in Figure~\ref{fig:bopp_overview}  do not contain runtime exception generators that disallow continued execution of programs that violate these constraints,  those actually implemented in the BOPP system do.

%% file: optimization.tex

In addition to the syntax introduced in the previous section, there are five main components to BOPP:
\begin{itemize}
	\setlength\itemsep{0.1em}
	\item[-] A program transformation, \texttt{q}$\rightarrow$\qmarg, allowing estimation of the evidence $p(Y,\theta)$ at a fixed $\theta$.
	\item[-] A high-performance, GP based, BO implementation for actively sampling $\theta$.
	\item[-] A program transformation, \texttt{q}$\rightarrow$\qprior,  used for automatic and adaptive domain scaling, such that a problem-independent hyperprior can be placed over the GP hyperparameters.
	\item[-] An adaptive non-stationary mean function to support unbounded optimization.
	\item[-] A program transformation, \texttt{q}$\rightarrow$\qacq, and annealing maximum likelihood estimation method to optimize the acquisition function subject the implicit constraints imposed by the generative model.
\end{itemize}
Together these allow BOPP to perform online MMAP estimation for arbitrary programs in a manner that is black-box from the user's perspective - requiring only the definition of the target program in the same way as existing PPS and identifying which variables to optimize.  The BO component of BOPP is both probabilistic programming and language independent, and is provided as a stand-alone package.\footnote{Code available at~\href{http://www.github.com/probprog/deodorant/}{\url{http://www.github.com/probprog/deodorant/}}}  It requires as input only a target function, a sampler to establish rough input scaling, and a problem specific optimizer for the acquisition function that imposes the problem constraints.  

\begin{figure}[t]
	\centering
	\includegraphics[width=\textwidth]{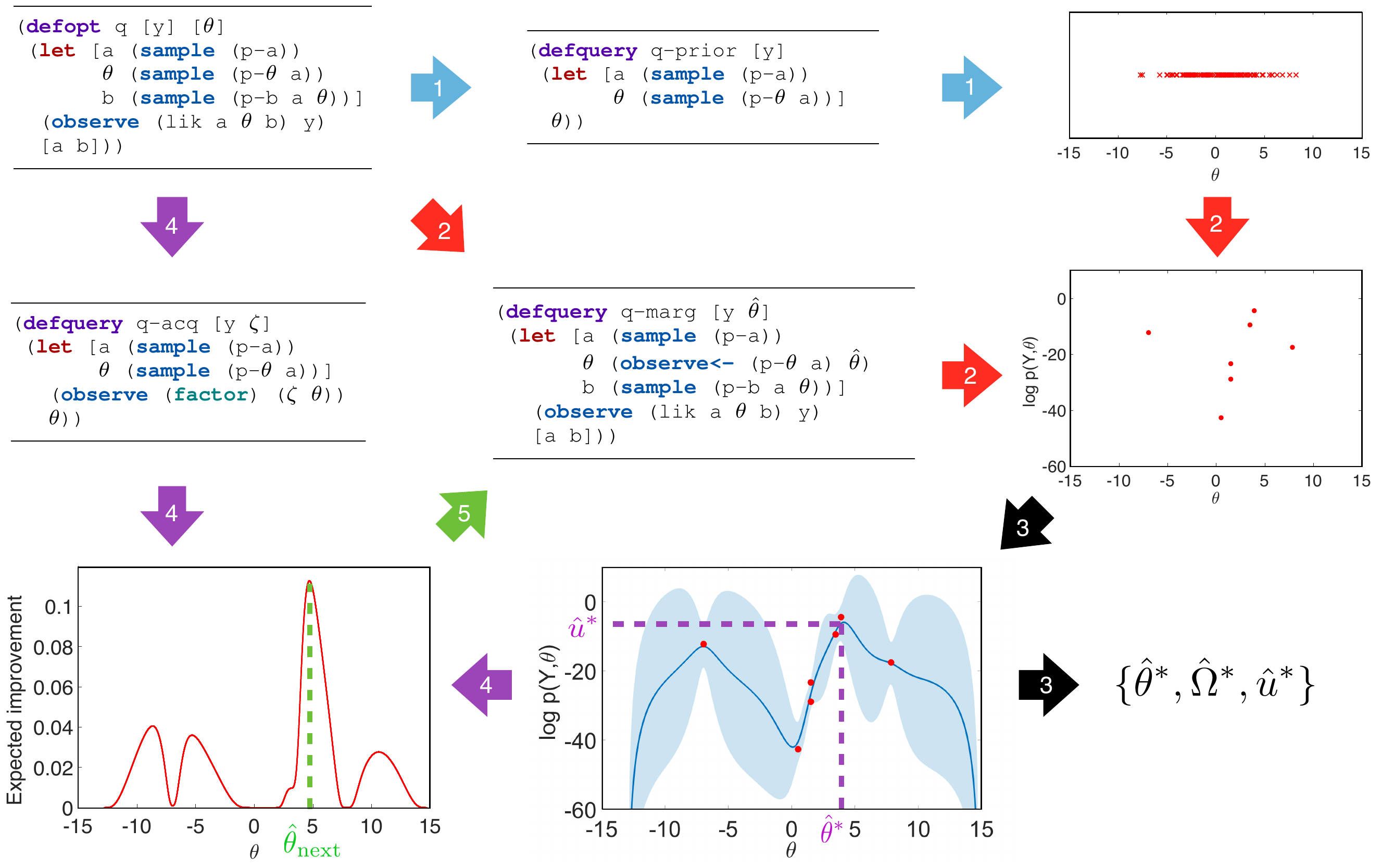}
	\caption{
		\label{fig:bopp_overview}
		Overview of the BOPP algorithm, description given in main text. \texttt{p-a}, \texttt{p-$\theta$}, \texttt{p-b} and \texttt{lik} all represent distribution object constructors. \factor is a special distribution constructor that assigns probability $p(y) = y$, in this case $y = \zeta(\theta)$.}
\end{figure}

Figure \ref{fig:bopp_overview} provides a high level overview of the algorithm invoked when \doopt is called on a query \texttt{q} that defines a distribution $p\left(Y, a, \theta , b\right)$.  We wish to optimize $\theta$ whilst marginalizing out $a$ and $b$, as indicated by the the second input to \texttt{q}. In summary, BOPP performs iterative optimization in 5 steps

\begin{itemize}
	\item[-] Step 1 (blue arrows) generates unweighted samples from the transformed prior program \texttt{q-prior} (\emph{top center}), constructed by removing all conditioning. This initializes the domain scaling for $\theta$.
	\item[-] Step 2 (red arrows) evaluates the marginal $p(Y,\theta)$ at a small number of the generated $\hth$ by performing inference on the marginal program \qmarg~ (\emph{middle centre}), which returns samples from the distribution $p\left(a,b | Y, \theta\right)$ along with an estimate of $p(Y, \theta)$.  The evaluated points (\emph{middle right}) provide an initial domain scaling of the outputs and starting points for the BO surrogate.
	\item[-] Step 3 (black arrow) fits a mixture of GPs posterior \cite{rasmussen2006gaussian} to the scaled data (\emph{bottom centre}) using a problem independent hyperprior. The solid blue line and shaded area show the posterior mean and $\pm2$ standard deviations respectively. The new estimate of the optimum $\hth^*$ is the value for which the mean estimate is largest, with $\hat{u}^*$ equal to the corresponding mean value.    
	\item[-] Step 4 (purple arrows) constructs an acquisition function $\zeta \colon \vartheta \rightarrow \real^+$ (\emph{bottom left}) using the GP posterior.  This is optimized, giving the next point to evaluate $\hth_{\mathrm{next}}$, by performing annealed importance sampling on a transformed program \texttt{q-acq} (\emph{middle left}) in which all \observe statements are removed and replaced with a single \observe assigning probability $\zeta(\theta)$ to the execution. 
	\item[-] Step 5 (green arrow) evaluates $\hth_{\mathrm{next}}$ using \qmarg~and continues to step 3.
\end{itemize}

\subsection{Program Transformation to Generate the Target}
\label{sec:transform}
\input{transform}

\subsection{Bayesian Optimization of the Marginal}
\label{sec:BOPP}

\input{bopp-for-ml}

\label{sec:bopp-for-ml}

\input{domainscaling}

\input{optacqfunc}

%% file: transform.tex

Consider the \defopt query \texttt{q} in Figure \ref{fig:bopp_overview}, the body of which defines the joint distribution $p\left(Y,a,\theta,b\right)$.   Calculating \eqref{eq:MMAP} (defining $X=\left\{a,b\right\}$) using a standard optimization scheme presents two issues: $\theta$ is a random variable within the program rather than something we control and its probability distribution is only defined conditioned on $a$.

We deal with both these issues simultaneously using a program transformation similar to the disintegration transformation in Hakaru \citep{zinkov2016composing}. Our \emph{marginal} transformation returns a new \query object, \qmarg~ as shown in Figure~\ref{fig:bopp_overview}, that defines the same joint distribution on program variables and inputs, but now accepts the value for $\theta$ as an input.  This is done by replacing all \sample statements associated with $\theta$ with equivalent \observes statements, taking $\theta$ as the observed value, where \observes is identical to \observe except that it returns the observed value.  As both \sample and \observe operate on the same variable type - a distribution object - this transformation can always be made, while the identical returns of \sample and \observes trivially ensures validity of the transformed program.

%% file: bopp-for-ml.tex

The target function for our BO scheme is $\log p(Y,\theta)$, noting $\argmax f\left(\theta\right) = \argmax \log f\left(\theta\right)$ for any $f : \vartheta \rightarrow \real^+$.  The log is taken because GPs have unbounded support, while $p\left(Y,\theta\right)$ is always positive, and because we expect variations over many orders of magnitude.  PPS with importance sampling based inference engines, e.g. sequential Monte Carlo \citep{wood2014new} or the particle cascade \citep{paige2014asynchronous}, can return noisy estimates of this target given the transformed program \qmarg.   



Our BO scheme uses a GP prior and a Gaussian likelihood.  Though the rationale for the latter is predominantly computational, giving an analytic posterior, there are also theoretical results suggesting that this choice is appropriate \citep{berard2014lognormal}. We use as a default covariance function a combination of a Mat\'{e}rn-3/2 and Mat\'{e}rn-5/2 kernel.  Specifically, let $D = \lVert \theta \rVert_0$ be the dimensionality of $\theta$ and define
\begin{subequations}
	\begin{align}
	\label{eq:d_def}
	d_{3/2}(\theta,\theta') &= \sqrt{\sum_{i=1}^{D} \frac{\theta_i-\theta_i'}{\rho_i}} \displaybreak[0] \\
	d_{5/2}(\theta,\theta') &= \sqrt{\sum_{i=1}^{D} \frac{\theta_i-\theta_i'}{\varrho_i}}
	\end{align}
\end{subequations}
where $i$ indexes a dimension of $\theta$ and $\rho_i$ and $\varrho_i$ are dimension specific length scale hyperparameters. Our prior covariance function is now given by
\begin{align}
\label{eq:kprior}
\begin{split}
k_{\text{prior}}\left(\theta,\theta'\right) = & \sigma_{3/2}^2 \left(1+\sqrt{3}d_{3/2}\left(\theta,\theta'\right)\right)\exp\left(-\sqrt{3}d_{3/2}\left(\theta,\theta'\right)\right) +\\&\sigma_{5/2}^2 \left(1+\sqrt{5}d_{5/2}\left(\theta,\theta'\right)+\frac{5}{3}(d_{5/2}\left(\theta,\theta'\right))^2\right)\exp\left(-\sqrt{5}d_{5/2}\left(\theta,\theta'\right)\right) 
\end{split}
\end{align}
where $\sigma_{3/2}$ and $\sigma_{5/2}$ represent signal standard deviations for the two respective kernels.  The full set of GP hyperparameters is defined by $\alpha = \{\sigma_n,\sigma_{3/2},\sigma_{5/2},\rho_{i=1:D},\varrho_{i=1:D}\}$.  A key feature of this kernel is that it is only once differentiable and therefore makes relatively weak assumptions about the smoothness of $f$.  The ability to include branching in a probabilistic program means that, in some cases, an even less smooth kernel than~\eqref{eq:kprior} might be preferable.  However, there is clear a trade-off between generality of the associated reproducing kernel Hilbert space and modelling power.

As noted by \citep{snoek2012practical}, the performance of BO using a single GP posterior is heavily influenced by the choice of these hyperparameters.  We therefore exploit the automated domain scaling introduced in Section~\ref{sec:domain} to define a problem independent hyperprior $p(\alpha)$ and perform inference to give a mixture of GPs posterior.  Details on this hyperprior are given in Appendix~\ref{sec:app:hyperprior}.

Inference over $\alpha$ is performed using Hamiltonian Monte Carlo (HMC) \citep{duane1987hybrid}, giving an unweighted mixture of GPs.  Each term in this mixture has an analytic distribution fully specified by its mean function $\mu_m^i \colon \vartheta \rightarrow \real$ and covariance function $k_m^i \colon \vartheta \times \vartheta \rightarrow \real$, where $m$ indexes the BO iteration and $i$ the hyperparameter sample.  HMC was chosen because of the availability of analytic derivatives of the GP log marginal likelihoods.  As we found that the performance of HMC was often poor unless a good initialization point was used, BOPP runs a small number of independent chains and allocates part of the computational budget to their initialization using a L-BFGS optimizer \citep{broyden1970convergence}. 

The inferred posterior is first used to estimate which of the previously evaluated $\hth_j$ is the most optimal, by taking the point with highest expected value
, $\hat{u}^*_m = \max_{j\in1\dots m} \sum_{i=1}^{N} \mu_{m}^i (\hth_j)$.  This completes the definition of the output sequence returned by the \doopt macro.  Note that as the posterior updates globally with each new observation, the relative estimated optimality of previously evaluated points changes at each iteration.
Secondly it is used to define the acquisition function $\zeta$, for which we take the expected improvement \citep{snoek2012practical}, defining $\sigma_m^i\left(\theta\right) = \sqrt{k_m^i\left(\theta,\theta\right)}$ and $\gamma_m^i\left(\theta\right) = \frac{\mu_m^i \left(\theta\right)-\hat{u}_m^*}{\sigma_m^i\left(\theta\right)}$,
\begin{align}
\label{eq:exp-imp}
\zeta \left(\theta\right) = \sum_{i=1}^{N} \left(\mu_m^i\left(\theta\right)-\hat{u}_m^*\right)\Phi \left(\gamma_m^i\left(\theta\right)\right)+\sigma_m^i\left(\theta\right)\phi\left(\gamma_m^i\left(\theta\right)\right)
\end{align}
where $\phi$ and $\Phi$ represent the pdf and cdf of a unit normal distribution respectively.   We note that more powerful, but more involved, acquisition functions, e.g. \citep{hernandez2014predictive}, could be used instead.

%% file: domainscaling.tex

\subsection{Automatic and Adaptive Domain Scaling}
\label{sec:domain}

Domain scaling, by mapping to a common space, is crucial for BOPP to operate in the required black-box fashion as it allows a general purpose and problem independent hyperprior to be placed on the GP hyperparameters.  BOPP therefore employs an affine scaling to a $[-1,1]$ hypercube for both the inputs and outputs of the GP.  To initialize scaling for the input variables, we sample directly from the generative model defined by the program. 
This is achieved using a second transformed program, \qprior, which removes all conditioning, i.e. \observe statements, and returns $\theta$.  This transformation also introduces code to terminate execution of the query once all $\theta$ are sampled, in order to avoid unnecessary computation.  As \observe statements return \lsi{nil}, this transformation trivially preserves the generative model of the program, but the probability of the execution changes.  Simulating from the generative model does not require inference or calling potentially expensive likelihood functions and is therefore computationally inexpensive.   By running inference on \qmarg~given a small number of these samples as arguments, a rough initial characterization of output scaling can also be achieved.  If points are observed that fall outside the hypercube under the initial scaling, the domain scaling is appropriately updated\footnote{An important exception is that the output mapping to the bottom of the hypercube remains fixed such that low likelihood new points are not incorporated.  This ensures stability when considering unbounded problems.} so that the target for the GP remains the $[-1,1]$ hypercube.


\subsection{Unbounded Bayesian Optimization via Non-Stationary Mean Function Adaptation}
\label{sec:unbounded}

Unlike standard BO implementations, BOPP is not provided with external constraints and we therefore develop a scheme for operating on targets with potentially unbounded support.  Our method exploits the knowledge that the target function is a probability density, implying that the area that must be searched in practice to find the optimum is finite, by defining a non-stationary prior mean function.  This takes the form of a bump function that is constant within a region of interest, but decays rapidly outside.  Specifically we define this bump function in the transformed space as
\begin{align}
\label{eq:BUMP}
\mu_{\mathrm{prior}}\left(r;r_e,r_{\mathrm{\infty}}\right) = \begin{cases} 0  \hfill & \mathrm{if} \; r \leq r_{\mathrm{e}} \\ 
\log \left(\frac{r-r_{\mathrm{e}}}{r_{\mathrm{\infty}}-r_{\mathrm{e}}}\right)+\frac{r-r_{\mathrm{e}}}{r_{\mathrm{\infty}}-r_{\mathrm{e}}} & \mathrm{otherwise} \end{cases}
\end{align}
where $r$ is the radius from the origin, $r_e$ is the maximum radius of any point generated in the initial scaling or subsequent evaluations, and $r_{\mathrm{\infty}}$ is a parameter set to $1.5 r_{e}$ by default.  Consequently, the acquisition function also decays and new points are never suggested arbitrarily far away.  Adaptation of the scaling will automatically update this mean function appropriately, learning a region of interest that matches that of the true problem, without complicating the optimization by over-extending this region.  We note that our method shares similarity with the recent work of Shahriari et al \citep{shahriari2016unbounded}, but overcomes the sensitivity of their method upon a user-specified bounding box representing soft constraints, by initializing automatically and adapting as more data is observed.

%% file: optacqfunc.tex

\subsection{Optimizing the Acquisition Function}
\label{sec:optacqfunc}

Optimizing the acquisition function for BOPP presents the issue that the query contains implicit constraints that are unknown to the surrogate function.  The problem of unknown constraints has been previously covered in the literature \citep{gardner2014bayesian,hernandez2015general} by assuming that constraints take the form of a black-box function which is modeled with a second surrogate function and must be evaluated in guess-and-check strategy to establish whether a point is valid. Along with the potentially significant expense such a method incurs, this approach is inappropriate for \emph{equality} constraints or when the target variables are potentially discrete.  For example, the Dirichlet distribution in Figure~\ref{fig:house-heating-code} introduces an equality constraint on \lsi{powers}, namely that its components must sum to $1$.

We therefore take an alternative approach based on directly using the program to optimize the acquisition function.  To do so we consider a transformed program \lsi{q-acq} that is identical to \lsi{q-prior} (see Section \ref{sec:domain}), but adds an additional \observe statement that assigns a weight $\zeta(\theta)$ to the execution.  By setting $\zeta(\theta)$ to the acquisition function, the maximum likelihood corresponds to the optimum of the acquisition function subject to the implicit program constraints.  
We obtain a maximum likelihood estimate for \lsi{q-acq} using a variant of annealed importance sampling \citep{neal2001annealed} in which lightweight Metropolis Hastings (LMH) \citep{wingate2011lightweight} with local random-walk moves is used as the base transition kernel. 


%% file: experiments.tex


\begin{figure*}[t]
	\includegraphics[width=0.99\textwidth]{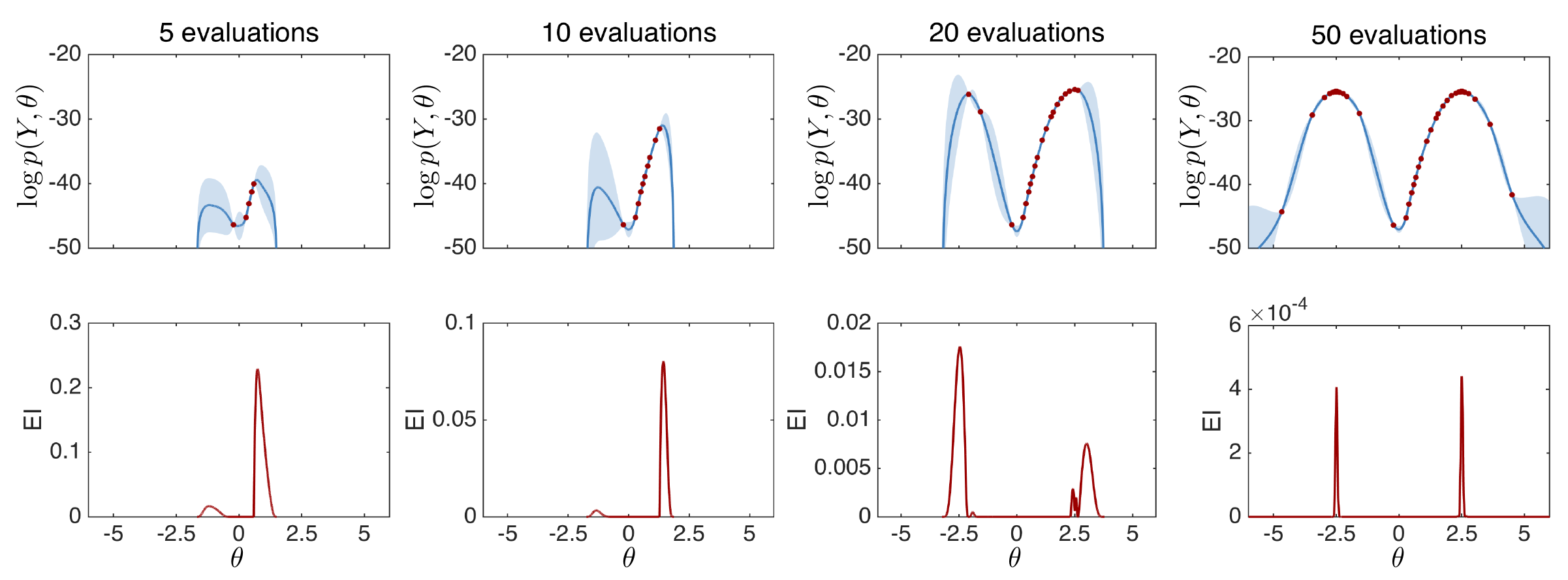}
	\caption{Convergence on an unconstrained bimodal problem with $p \left(\theta\right)={\rm Normal}(0, 0.5)$ and $p \left(Y|\theta\right)={\rm Normal}(5-\left|\theta\right|,0.5)$ giving significant prior misspecification. The top plots show a regressed GP, with the solid line corresponding to the mean and the shading shows $\pm$ 2 standard deviations.  The bottom plots show the corresponding acquisition functions. \label{fig:domainAdpat}}
\end{figure*}

We first demonstrate the ability of BOPP to carry out unbounded optimization using a 1D problem with a significant prior-posterior mismatch as shown in Figure \ref{fig:domainAdpat}.  It shows BOPP adapting to the target and effectively establishing a maxima in the presence of multiple modes.   After 20 evaluations the acquisitions begin to explore the right mode, after 50 both modes have been fully uncovered.

\begin{figure*}[t]
	\centering
	\includegraphics[width=1\textwidth]{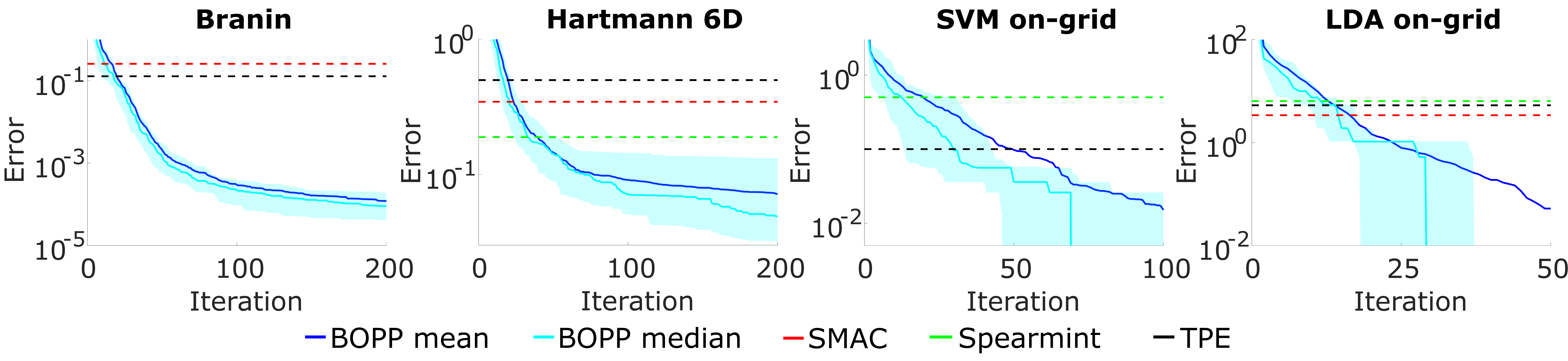}
	\caption{Comparison of BOPP  used as an optimizer to prominent BO packages on common benchmark problems.  
		The dashed lines shows the final mean error of SMAC (red), Spearmint (green) and TPE (black) as quoted by \cite{eggensperger2013towards}. 
		The dark blue line shows the mean error for BOPP averaged over 100 runs, whilst the median and 25/75\% percentiles are shown in cyan. Results for Spearmint on Branin and SMAC on SVM on-grid are omitted because both BOPP and the respective algorithms averaged zero error to the provided number of significant figures in \cite{eggensperger2013towards}.
		\label{fig:bayes-opt}}
\end{figure*}

\subsection{Classic Optimization Benchmarks}

Next we compare BOPP to the prominent BO packages SMAC \cite{hutter2011sequential}, Spearmint \cite{snoek2012practical} and TPE \cite{bergstra2011algorithms} on a number of classical benchmarks as shown in Figure \ref{fig:bayes-opt}.  These results demonstrate that BOPP provides substantial advantages over these systems when used simply as an optimizer on both continuous and discrete optimization problems.  In particular, it offers a large advantage over SMAC and TPE on the continuous problems (Branin and Hartmann), due to using a more powerful surrogate, and over Spearmint on the others due to not needing to make approximations to deal with discrete problems.

\subsection{Marginal Maximum a Posteriori Estimation Problems}

We now demonstrate application of BOPP on a number of MMAP problems.  Comparisons here are more difficult due to the dearth of existing alternatives for PPS.  In particular, simply running inference on the original query does not return estimates for $p\left(Y,\theta\right)$.  We consider the possible alternative of using our conditional code transformation to design a particle marginal Metropolis Hastings (PMMH, \cite{andrieu2010particle}) sampler which operates in a similar fashion to BOPP except that new $\theta$ are chosen using a MH step instead of actively sampling with BO.
For these MH steps we consider both LMH \citep{wingate2011lightweight} with proposals from the prior and the random-walk MH (RMH) variant introduced in Section \ref{sec:optacqfunc}.

\subsubsection{Hyperparameter Optimization for Gaussian Mixture Model}

\input{mvn-mixture.tex}

\subsubsection{Extended Kalman Filter for the Pickover Chaotic Attractor}
\label{sec:AppKalman}

\input{chaos.tex}

\subsubsection{Hidden Markov Model with Unknown Number of States}

\input{hmm.tex}

%% file: mvn-mixture.tex

\begin{figure}[t]
	\begin{lstlisting}[basicstyle=\footnotesize\ttfamily]
(defopt mvn-mixture [data mu0 kappa psi] [nu alpha]
 (let [[n d] (shape data)
       alpha (sample (uniform-continuous 0.01 100))
       nu (sample (uniform-continuous (- d 1) 100))
       obs-proc0 (mvn-niw mu0 kappa nu psi)]
       (loop [data data
              obs-procs {}
              mix-proc (dirichlet-discrete 
                          (vec (repeat d alpha)))]
	    (let [y (first data)]
	     (if y
	      (let [z (sample (produce comp-proc))
	            obs-proc (get obs-procs z obs-proc0)
	            obs-dist (produce obs-proc)]
	        (observe obs-dist y)
	        (recur (rest data)
	               (assoc obs-procs z (absorb obs-proc y))
	        (absorb mix-proc z)))
	      mix-proc)))))
	\end{lstlisting}
	\caption{
		\label{fig:mvn-code}
		Anglican query for hyperparameter optimization of a Gaussian mixture model, defined in terms of two parameters \lsi{nu} and \lsi{alpha}. A \lsi{mvn-niw} process is used to represent the marginal likelihood of observations under a Gaussian-inverse-Wishart prior, whereas a \lsi{dirichlet-discrete} process models the prior probability of cluster assignments under a Dirichlet-discrete prior. The command \lsi{produce} returns the predictive distribution for the next sample from a process. \lsi{absorb} conditions on the value of the next sample.}
\end{figure}

\begin{figure*}[t]
	\includegraphics[width=\textwidth]{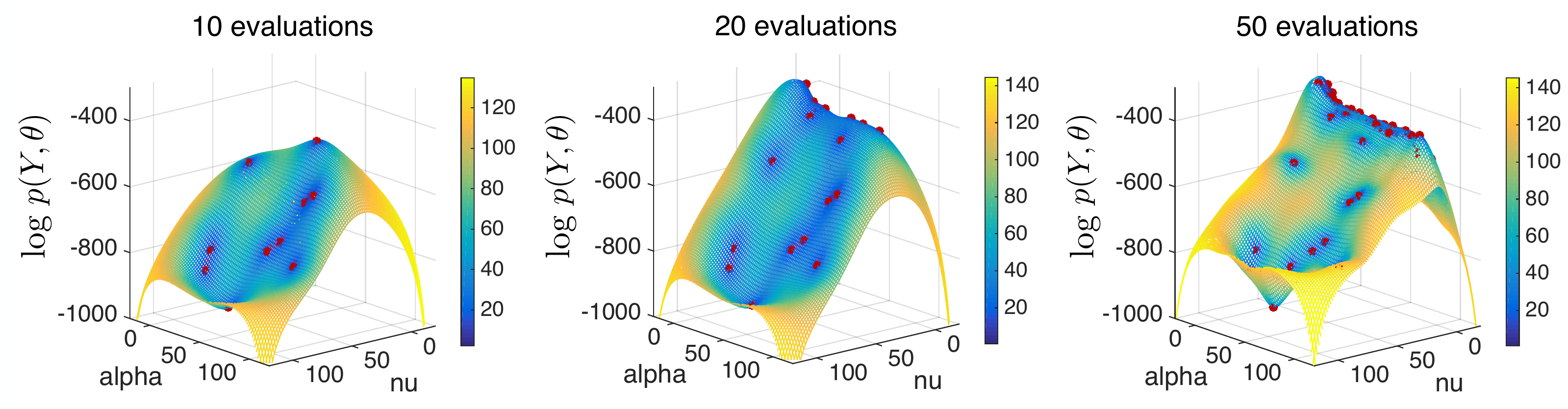}
	\caption{
		\label{fig:mvn-gp-surface}
		Bayesian optimization of hyperparameters in a Gaussian mixture model evaluated on the Iris dataset. Panels show the GP posterior as a function of number of evaluations, with the surface corresponding to the posterior mean and the color bars the posterior standard deviation. Optimization is performed over the parameter $\alpha$ of a 10-dimensional symmetric Dirichlet distribution and the degrees of freedom $\nu$ of the inverse-Wishart prior. At each evaluation we obtain an estimate of the log marginal $\log p(Y,\theta)$ obtained by performing sequential Monte Carlo inference with 1000 particles.  The apparent maximum after initialization with 10 randomly sampled points lies at $\nu=31$, $\alpha=60$, and $\log p(Y,\theta) = -456.3$ (\emph{left}).  The surface after 10 optimization steps shows a new maximum at $\nu=9.2$, $\alpha=0.8$, and $\log p(Y,\theta) = -364.2$ (\emph{middle}). After 40 steps and 50 total evaluations this optimum is refined to $\nu=16$, $\alpha = 0.2$, and $\log p(Y,\theta) = -352.5$  (\emph{right}).}
\end{figure*}

We start with an illustrative case study of optimizing the hyperparameters in a multivariate Gaussian mixture model. We consider a Bayesian formulation with a symmetric Dirichlet prior on the mixture weights and a Gaussian-inverse-Wishart prior on the likelihood parameters:
\begin{align}
\v{\pi}
&\sim 
{\rm Dir(\alpha, \ldots, \alpha)}
\displaybreak[0]\\
(\v{\mu}_k, \v{\Sigma}_k)
&\sim 
{\rm NIW} (\v{\mu}_0, \kappa, \v{\Psi}, \nu)
&
{\rm for~}
&
k = 1, \ldots , K
\displaybreak[0]\\
z_n 
&\sim 
{\rm Disc(\v{\pi})}
\displaybreak[0]\\
\v{y}_n
&
\sim
{\rm Norm}(\v{\mu}_{z_n}, \v{\Sigma}_{z_n})
&
{\rm for~}
&
n = 1, \ldots , N
\end{align}
Anglican code for this model is shown in Figure 4. Anglican provides stateful objects, which are referred to as random processes, to represent the predictive distributions for the cluster assignments $z$ and the observations $\v{y}^k$ assigned to each cluster
\begin{align}
z_{n+1}
& \sim 
p( \cdot \,|\, z_{1:n}, \alpha),
\\
\v{y}_{m+1}^{k} 
& \sim 
p(\cdot \,|\, \v{y}^k_{1:m}, \v{\mu}_0, \kappa, \v{\Psi}, \nu).
\end{align}
In this collapsed representation marginalization over the model parameters $\v{\pi}$, $\v{\mu}_{k=1:K}$, and $\v{\Sigma}_{k=1:K}$ is performed analytically.
Using the Iris dataset, a standard benchmark for mixture models that contains 150 labeled examples with 4 real-valued features, we optimize the marginal with respect to the subset of the parameters $\nu$ and $\alpha$ under uniform priors over a fixed interval.  For this model, BOPP aims to maximize
\begin{align}
\begin{split}
& p(\nu, \alpha | \v{y}_{n=1:N}, \v{\mu}_0, \kappa, \v{\Psi}) \\
&= \iiiint p(\nu, \alpha, z_{n=1:N}, \v{\pi}, \v{\mu}_{k=1:K}, \v{\Sigma}_{k=1:K} | \v{y}_{n=1:N}, \mu_0, \kappa, \v{\Psi}) \mathrm{d}z_{n=1:N}\mathrm{d}\v{\pi}\mathrm{d}\v{\mu}_{k=1:K}\mathrm{d}\v{\Sigma}_{k=1:K}.
\end{split}
\end{align}

Figure~\ref{fig:mvn-gp-surface} shows GP regressions on the evidence after different numbers of the SMC evaluations have been performed on the model.  This demonstrates how the GP surrogate used by BO builds up a model of the target, used to both estimate the expected value of $\log p(Y,\theta)$ for a particular $\theta$ and actively sample the $\theta$ at which to undertake inference.

%% file: chaos.tex

\begin{figure*}[p]
	\centering
	\includegraphics[width=2.72in]{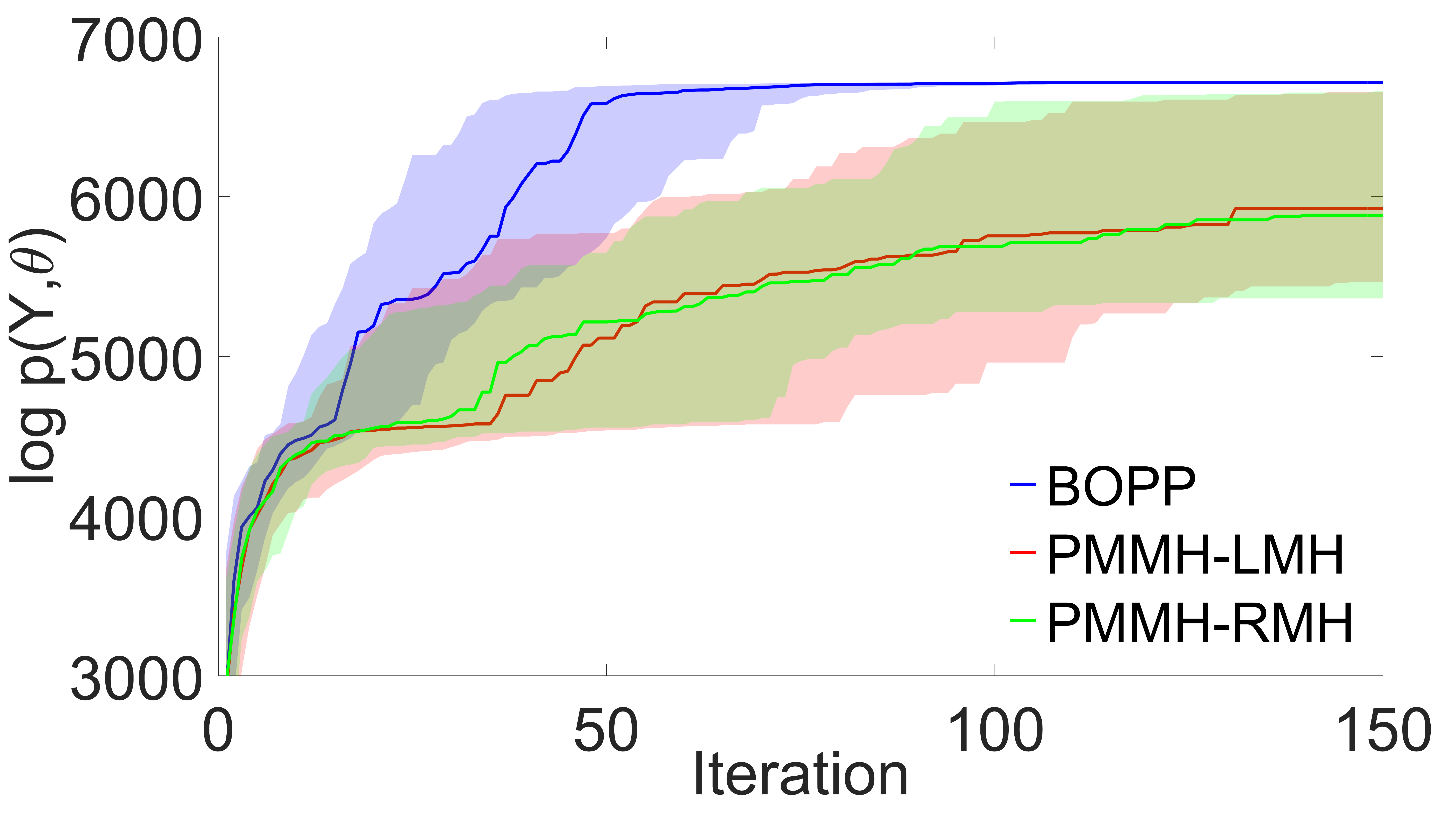}
	\includegraphics[width=2.72in]{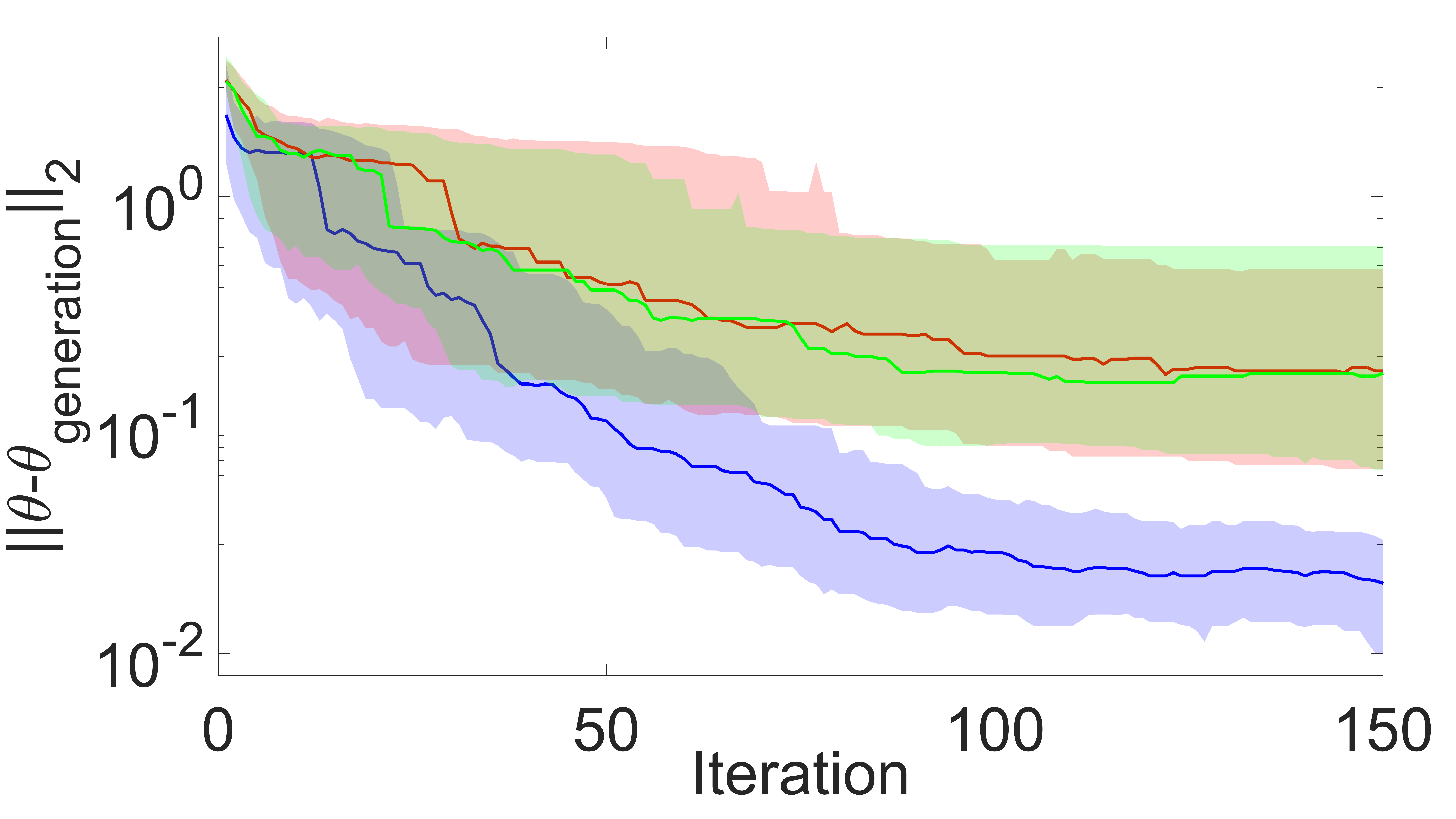}
	\caption{Convergence for transition dynamics parameters of the pickover attractor in terms of the cumulative best $\log p\left(Y,\theta\right)$ (\emph{left}) and distance to the ``true" $\theta$ used in generating the data (\emph{right}). Solid line shows median over 100 runs, whilst the shaded region the 25/75\% quantiles.  \label{fig:chaos}
		\vspace{6pt}}
\end{figure*}

\begin{figure}[p]
	\centering
	\begin{subfigure}[t]{0.48\textwidth}
		\centering
		\includegraphics[height=5.6cm,width=6.4cm]{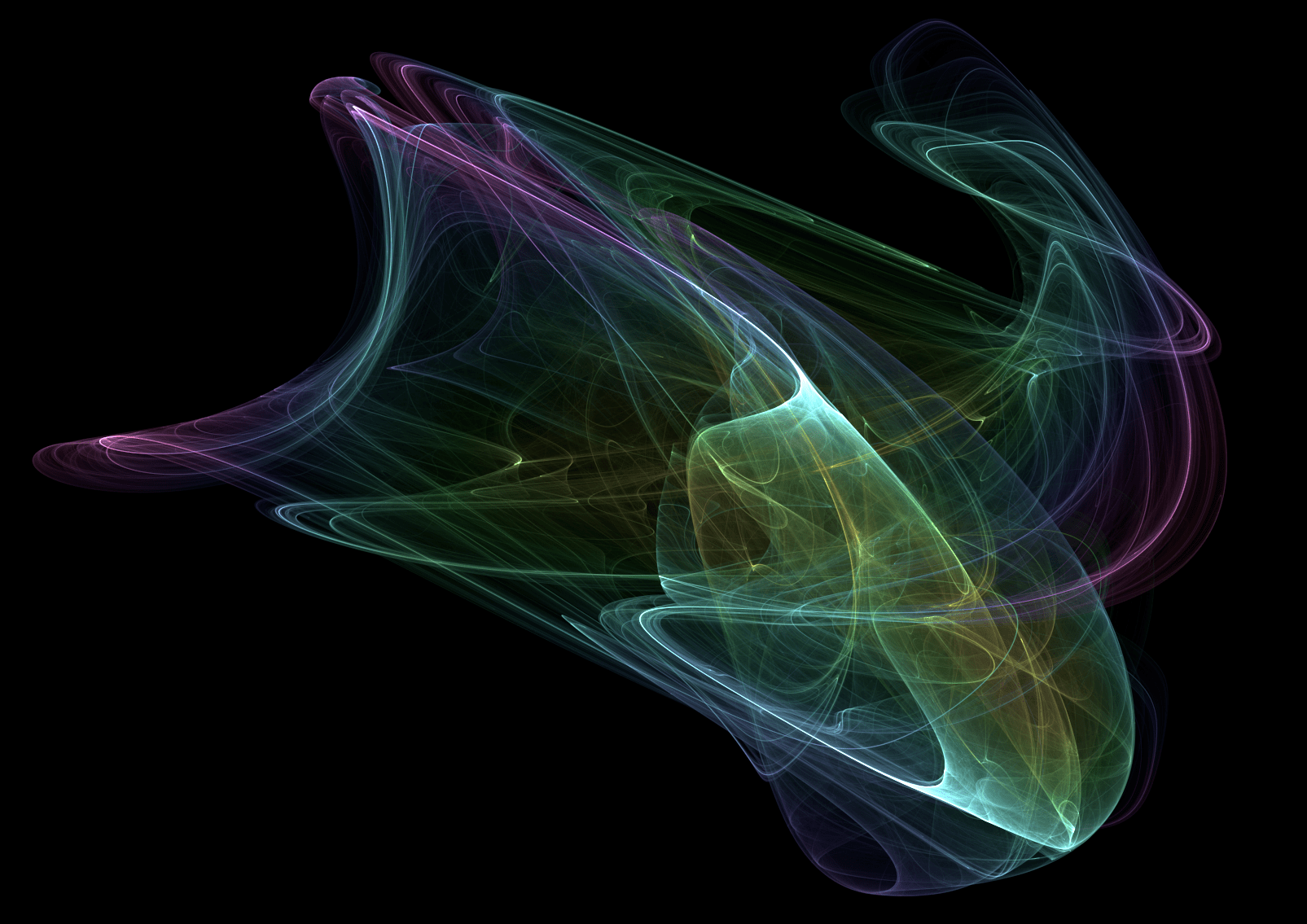}
		\caption{{$\begin{array}{c}
				\text{1 iteration}, \; 
				\theta = [-1.478,0.855]^T 
				\end{array}$}}
	\end{subfigure}
	\begin{subfigure}[t]{0.48\textwidth}
		\centering
		\tiny
		\includegraphics[height=5.6cm,width=6.4cm]{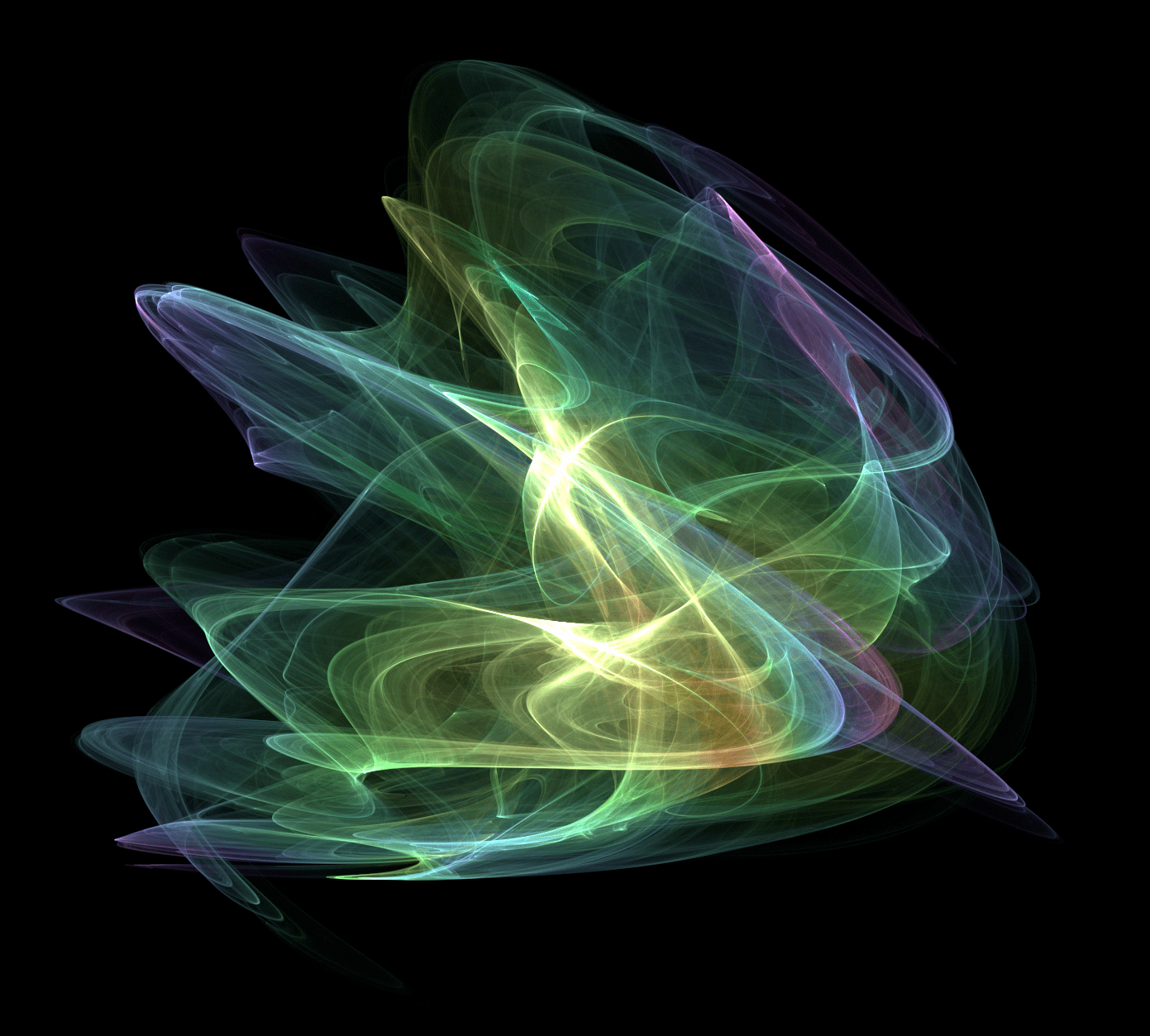}
		\caption{{$\begin{array}{c}
				\text{20 iterations}, \; 
				\theta = [-2.942,1.550]^T 
				\end{array}$}}
	\end{subfigure}
	\\
	\vspace{20pt}
	\begin{subfigure}[t]{0.48\textwidth}
		\centering
		\tiny
		\includegraphics[height=5.6cm,width=6.4cm]{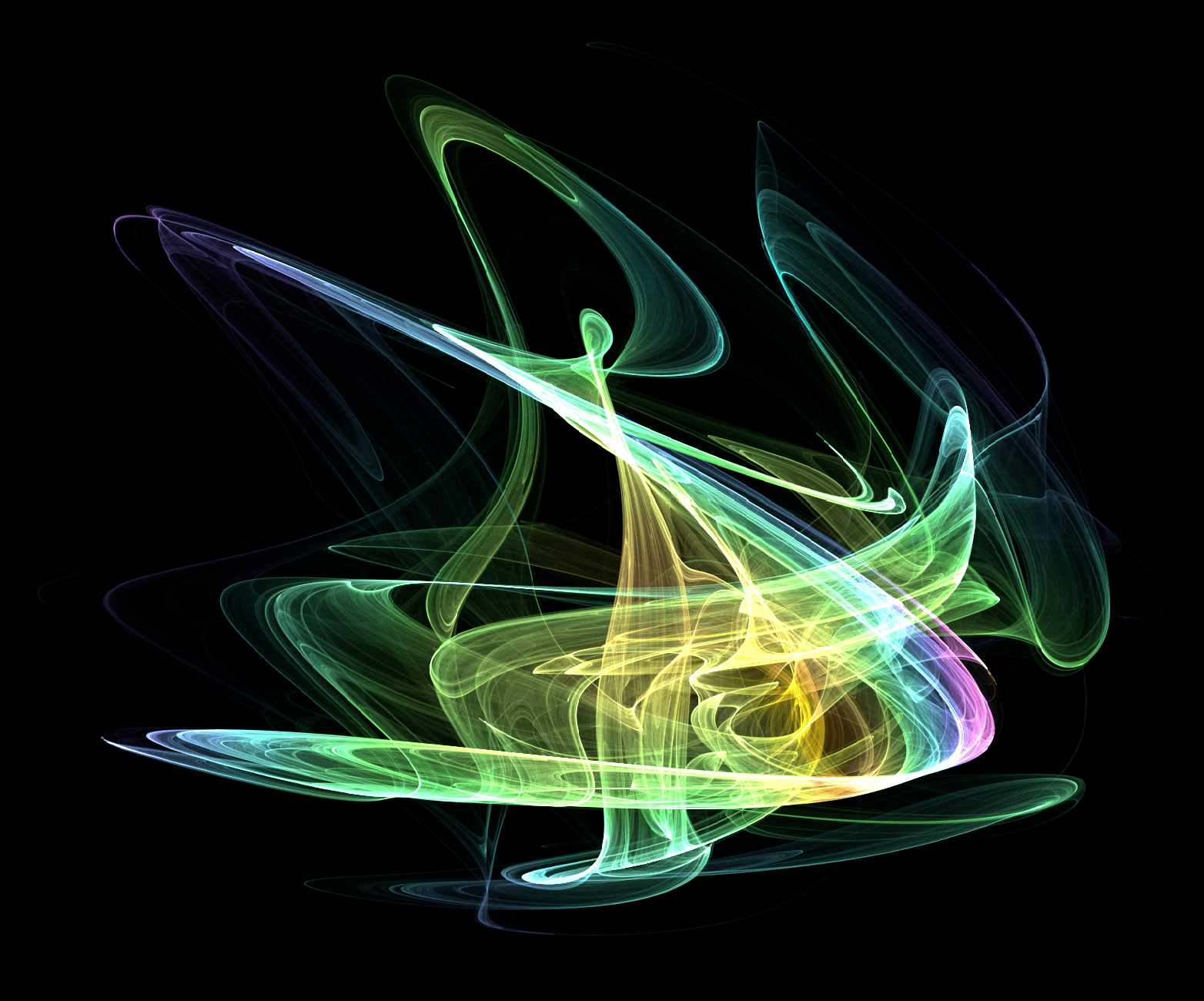}
		\caption{{$\begin{array}{c}
				\text{100 iterations}, \;
				\theta = [-2.306,1.249]^T 
				\end{array}$}}
	\end{subfigure}
	\begin{subfigure}[t]{0.48\textwidth}
		\centering
		\tiny
		\includegraphics[height=5.6cm,width=6.4cm]{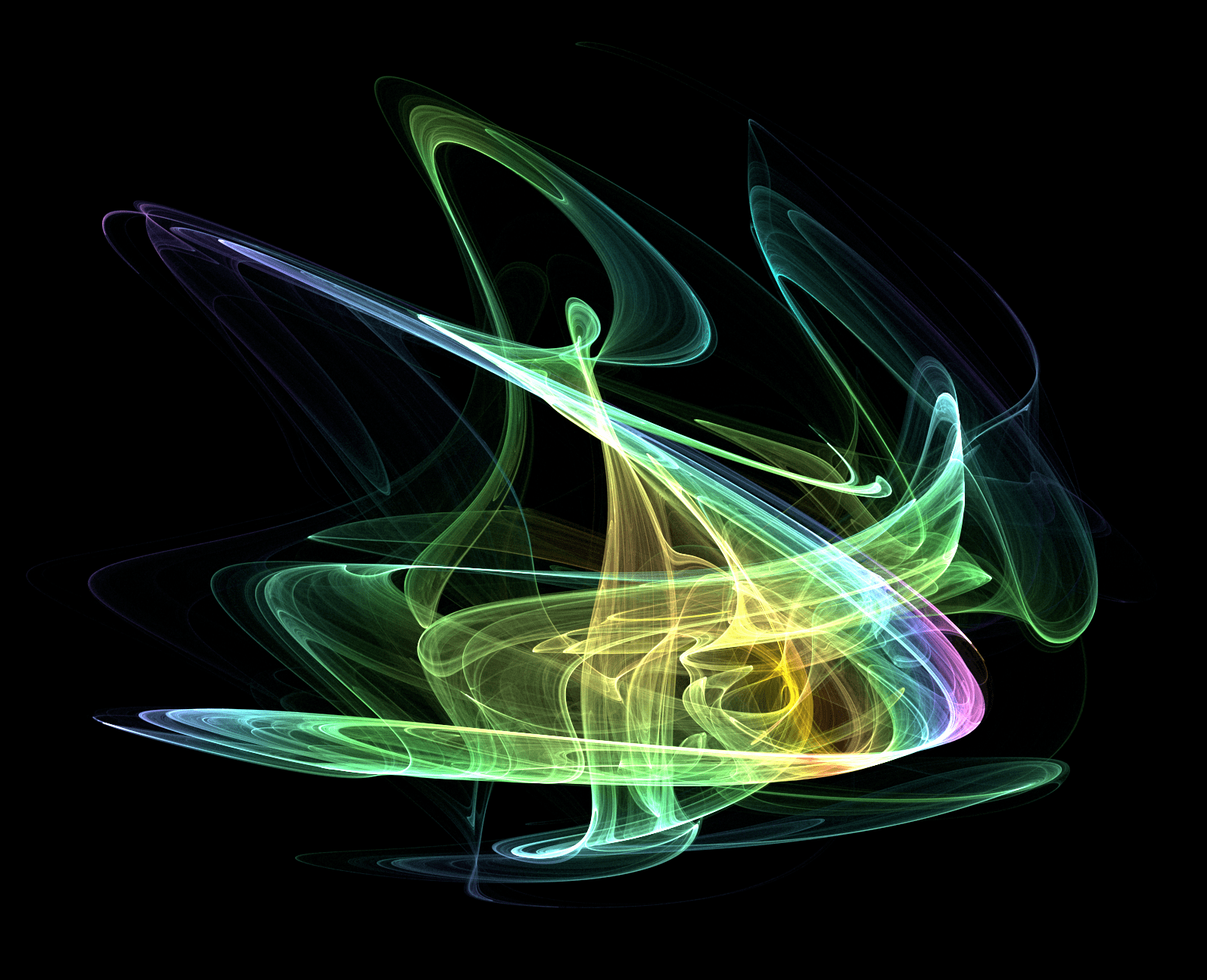}
		\caption{{$\begin{array}{c}
				\text{Ground truth}, \;
				\theta = [-2.3,1.25]^T
				\end{array}$}}
	\end{subfigure}
	\caption{A series of trajectories for different parameters, demonstrating convergence to the true attractor.  The colormap is based on the speed and curvature of the trajectory, with rendering done using the program Chaoscope (available at {\href{http://www.chaoscope.org/}{http://www.chaoscope.org/}}). \label{fig:chaoscope}}
\end{figure}

We next consider the case of learning the dynamics parameters of a chaotic attractor.  Chaotic attractors present an interesting case for tracking problems as, although their underlying dynamics are strictly deterministic with bounded trajectories, neighbouring trajectories diverge exponentially\footnote{It is beyond the scope of this paper to properly introduce chaotic systems.  We refer the reader to \cite{devaney1989introduction} for an introduction.}.  Therefore regardless of the available precision, a trajectory cannot be indefinitely extrapolated to within a given accuracy and probabilistic methods such as the extended Kalman filter must be incorporated \citep{fujii2013extended,ruan2003chaotic}. From an empirical perspective, this forms a challenging optimization problem as the target transpires to be multi-modal, has variations at different length scales, and has local minima close to the global maximum.

Suppose we observe a noisy signal $y_t \in \real^{K}, \; t = 1,2,\dots,T$ in some $K$ dimensional observation space were each observation has a lower dimensional latent parameter $x_t \in \real^{D},  \; t = 1,2,\dots,T$ whose dynamics correspond to a chaotic attractor of known type, but with unknown parameters.  Our aim will be to find the MMAP values for the dynamics parameters $\theta$, marginalizing out the latent states.  The established parameters can then be used for forward simulation or tracking.  

To carry out the required MMAP estimation, we apply BOPP to the extended Kalman smoother
\begin{align}
\label{eq:Kalman}
x_1 \sim & \mathcal{N} \left(\mu_1, \sigma_1 I\right) \\
x_t = & A \left(x_{t-1}, \theta\right)+\delta_{t-1}, \; & \delta_{t-1} \sim \mathcal{N} \left(0, \sigma_q I\right) \\
y_t = & C x_{t}+\varepsilon_{t}, \; & \varepsilon_{t} \sim \mathcal{N} \left(0, \sigma_y I\right)
\end{align}
where $I$ is the identity matrix, $C$ is a known $K \times D$ matrix,  $\mu_1$ is the expected starting position, and $\sigma_1, \sigma_q$ and $\sigma_y$ are all scalars which are assumed to be known.  The transition function $A \left(\cdot,\cdot\right)$
\begin{subequations}
	\begin{align}
	\label{eq:pickover}
	x_{t,1} = & \sin \left(\beta x_{t-1,2}\right)-\cos\left(\frac{5x_{t-1,1}}{2}\right)x_{t-1,3}  \\
	x_{t,2} = & -\sin \left(\frac{3x_{t-1,1}}{2}\right)x_{t-1,3}-\cos\left(\eta x_{t-1,2}\right) \\
	x_{t,3} = & \sin \left(x_{t-1,1}\right)
	\end{align}
\end{subequations}
corresponds to a Pickover attractor \citep{pickover1995pattern} with unknown parameters $\theta = \left\{\beta,\eta\right\}$ which we wish to optimize.  Note that $\eta$ and $-\eta$ will give the same behaviour.

Synthetic data was generated for $500$ time steps using the parameters of  $\mu_1 = [-0.2149,-0.0177,0.7630]^T$, $\sigma_1 = 0$, $\sigma_q = 0.01$, $\sigma_y = 0.2$, a fixed matrix $C$ where $K=20$ and each column was randomly drawn from a symmetric Dirichlet distribution with parameter $0.1$, and ground truth transition parameters of $\beta = -2.3$ and $\eta = 1.25$ (note that the true global optimum for finite data need not be exactly equal to this).  

MMAP estimation was performed on this data using the same model and parameters, with the exceptions of $\theta$, $\mu_1$ and $\sigma_1$.  
The prior on $\theta$ was set to a uniform in over a bounded region such that
\begin{align}
\label{eq:priorKalman}
p\left(\beta,\eta\right) = \begin{cases}
1/18, & \mathrm{if} -3 \le \beta \le 3 \; \cap \; 0 \le \eta \le 3\\
0, & \mathrm{otherwise}
\end{cases}.
\end{align}
The changes $\mu_1 = [0,0,0]$ and $\sigma_1 = 1$ were further made to reflect the starting point of the latent state being unknown.   For this problem, BOPP aims to maximize
\begin{align}
p(\beta,\eta | y_{t=1:T}) = \int p(\beta,\eta,x_{t=1:T} | y_{t=1:T}) \mathrm{d}x_{t=1:T} .
\end{align}
Inference on the transformed marginal query was carried out using SMC with 500 particles.  Convergence results are given in Figure~\ref{fig:chaos} showing that BOPP comfortably outperforms the PMMH variants, while Figure~\ref{fig:chaoscope} shows the simulated attractors generated from the dynamics parameters output by various iterations of a particular run of BOPP.

%% file: hmm.tex

\begin{figure*}[t]
	\centering
	\includegraphics[width=2.72in]{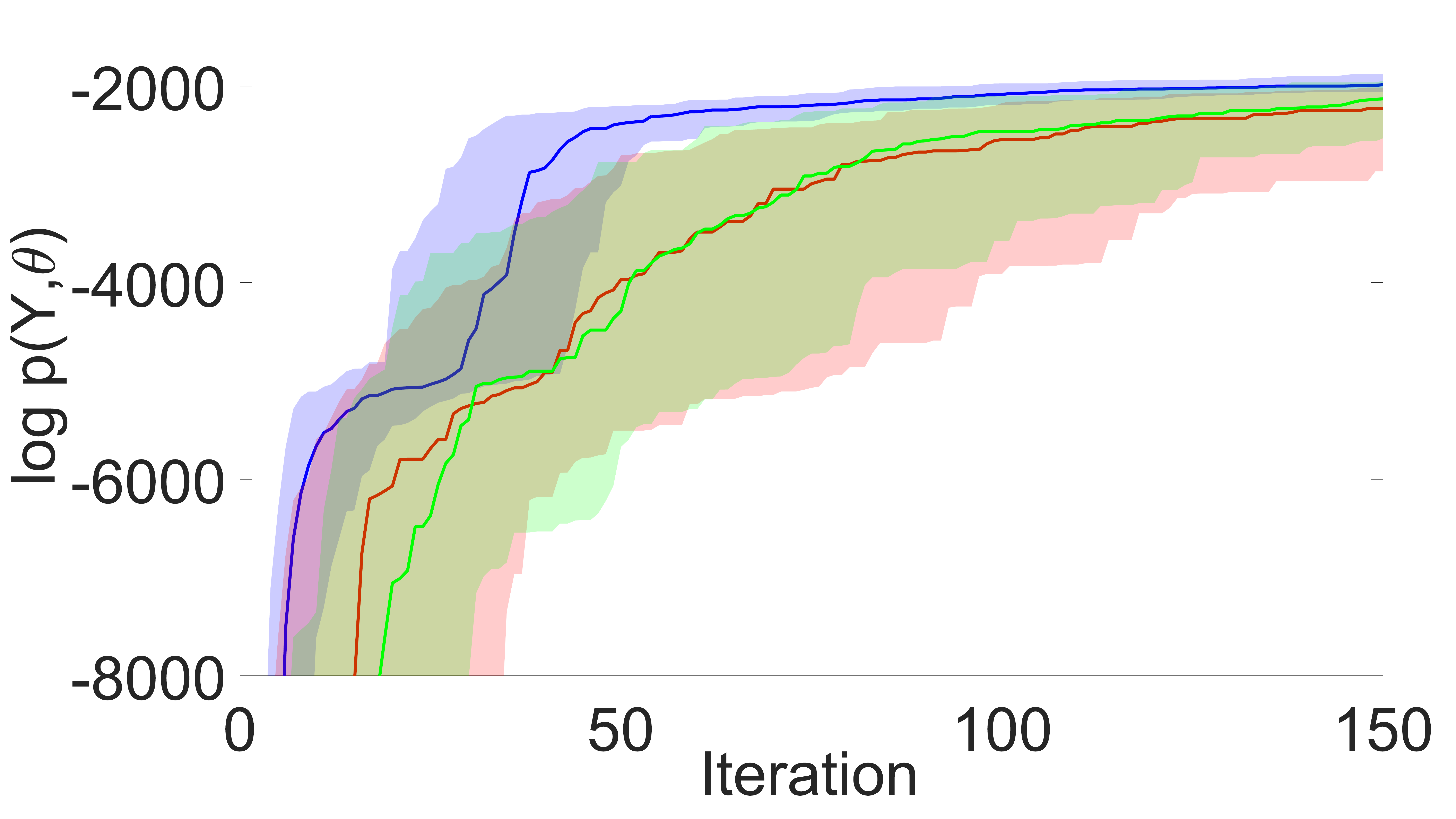}
	\includegraphics[width=2.72in]{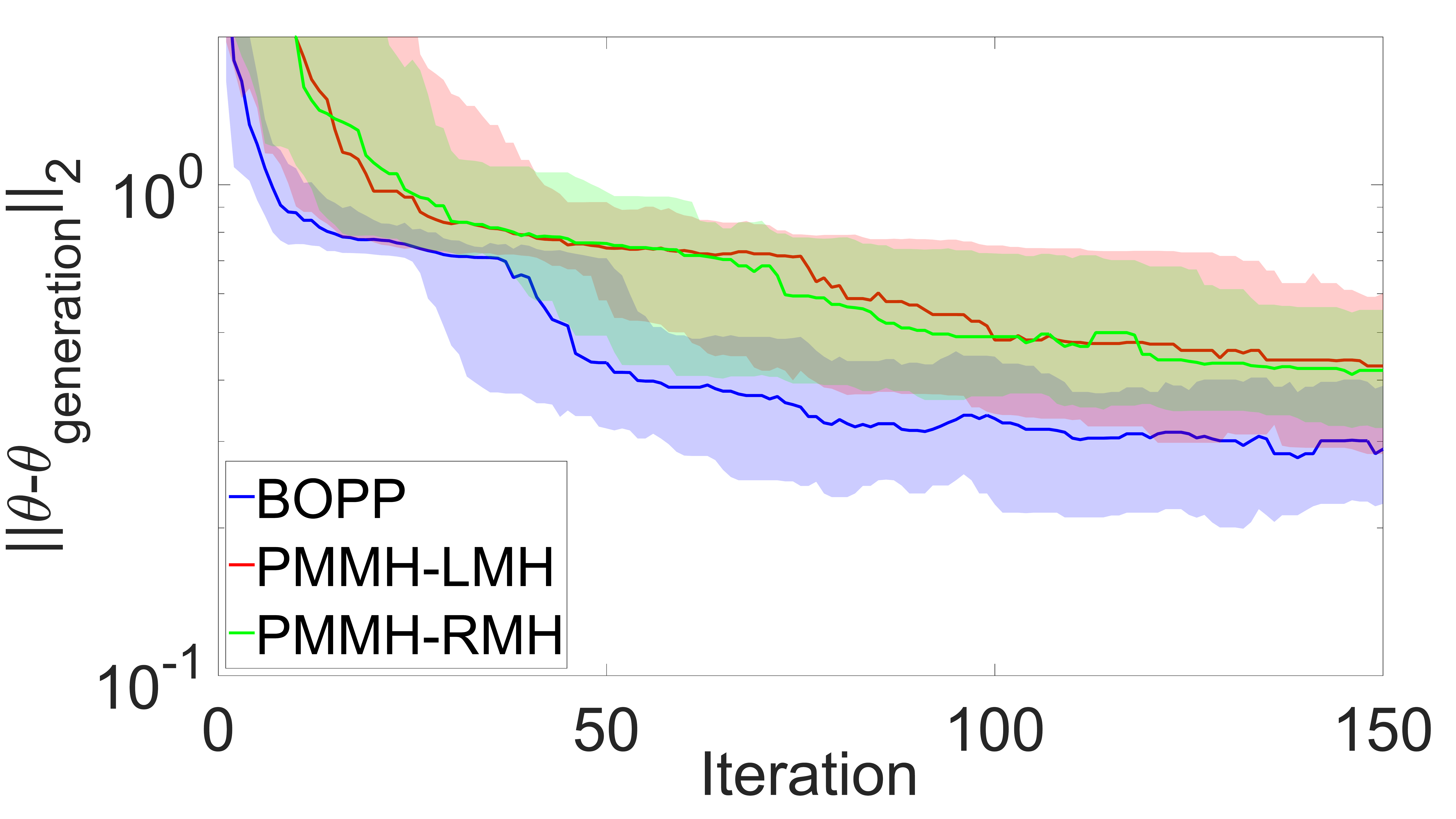}
	\caption{Convergence for HMM in terms of the cumulative best $\log p\left(Y,\theta\right)$ (\emph{left}) and distance to the ``true" $\theta$ used in generating the data (\emph{right}). Solid line shows median over 100 runs, whilst the shaded region the 25/75\% quantiles.  Note that for the distance to true $\theta$ was calculated by selecting which three states (out of the 5 generates) that were closest to the true parameters.  \label{fig:hmm}}
\end{figure*}

We finally consider a hidden Markov model (HMM) with an unknown number of states.  This example demonstrates how BOPP can be applied to models which conceptually have an unknown number of variables, by generating all possible variables that might be needed, but then leaving some variables unused for some execution traces.  This avoids problems of varying base measures so that the MMAP problem is well defined  and provides a function with a fixed number of inputs as required by the BO scheme.  From the BO perspective, the target function is simply constant for variations in an unused variable.

HMMs are Markovian state space models with discrete latent variables.  Each latent state $x_t \in\{1,\dots,K\}, t=1,\dots,T$ is defined conditionally on $x_{t-1}$ through a set of discrete transition probabilities, whilst each output $y_t\in\real$ is considered to be generated i.i.d. given $x_t$.  We consider the following HMM, in which the number of states $K$, is also a random variable: 
\begin{align}
\label{eq:hmm}
K & \sim \text{Discrete}\{1,2,3,4,5\} \\\displaybreak[0]
T_k &\sim \text{Dirichlet}\{{1}_{1:K}\}, \quad \forall k=1,\dots,K \\\displaybreak[0]
\phi_k &\sim \text{Uniform}[0,1], \quad \forall k=1,\dots,K \\\displaybreak[0]
\mu_0 &\leftarrow \min \{y_{1:T}\} \\\displaybreak[0]
\mu_k &\leftarrow \mu_{k-1}+\phi_k \cdot (\max \{y_{1:T}\} -\mu_{k-1}), \quad \forall k=1,\dots,K \\\displaybreak[0]
x_1 &\leftarrow 1 \\\displaybreak[0]
x_t | x_{t-1} &\sim \text{Discrete}\{T_{x_{t-1}}\}\\\displaybreak[0]
y_t | x_t &\sim\mathcal{N}(\mu(x_{t-1}),0.2).
\end{align}
Our experiment is based on applying BOPP to the above model to do MMAP estimation with a single synthetic dataset, generated using $K=3, \;\mu_1 = -1, \;\mu_2 = 0, \;\mu_3 = 4, \;T_1 = [0.9,0.1,0], \;T_2=[0.2,0.75,0.05]$ and $T_3=[0.1,0.2,0.7]$.  

We use BOPP to optimize both the number of states $K$ and the stick-breaking parameters $\phi_k$, with full inference performed on the other parameters.  BOPP therefore aims to maximize
\begin{align}
\label{eq:hmm-marginal}
p(K,\phi_{k=1:5}|y_{t=1:T}) = \iint p(K,\phi_{k=1:5},x_{t=1:T},T_{k=1:K}|y_{t=1:T}) \mathrm{d}x_{t=1:T} \mathrm{d}T_{k=1:K}.
\end{align}
As with the chaotic Kalman filter example, we compare to two PMMH variants using the same code transformations.  The results, given in Figure~\ref{fig:hmm}, again show that BOPP outperforms these PMMH alternatives.

%% file: discussion.tex

We have introduced a new method for carrying out MMAP estimation of probabilistic program variables using Bayesian optimization, representing the first unified framework for optimization and inference of probabilistic programs.  By using a series of code transformations, our method allows an arbitrary program to be optimized with respect to a defined subset of its variables, whilst marginalizing out the rest.  To carry out the required optimization, we introduce a new GP-based BO package that exploits the availability of the target source code to provide a number of novel features, such as automatic domain scaling and constraint satisfaction.  

The concepts we introduce lead directly to a number of extensions of interest, including but not restricted to smart initialization of inference algorithms, adaptive proposals, and nested optimization.  Further work might consider maximum marginal likelihood estimation and risk minimization.  Though only requiring minor algorithmic changes, these cases require distinct theoretical considerations.


%% file: program-transformations.tex

In this section we give a more detailed and language specific description of our program transformations, code for which can be found at \href{http://www.github.com/probprog/bopp}{\url{http://www.github.com/probprog/bopp}}. 

\subsection{Anglican}
Anglican is a probabilistic programming language integrated into Clojure (a dialect of Lisp) and inherits most of the corresponding syntax. Anglican extends Clojure with the special forms \sample and \observe \citep{tolpin2015probabilistic}.  
Each random draw in an Anglican program corresponds to a \sample  call, which can be thought of as a term in the prior. 
Each \observe statement applies weighting to a program trace and thus constitutes a term in the likelihood.
Compilation of an Anglican program, performed by the macro \lsi{query}, corresponds to transforming the code into a variant of continuation-passing style (CPS) code, which results in a function that can be executed using a particular inference algorithm.

Anglican program code is represented by a nested list of expressions, symbols, non-literals for contructing data structures (e.g. \lsi{[...]} for vectors), and command dependent literals (e.g. \lsi{[...]} as a second argument of a \lsi{let} statement which is used for binding pairs).  In order to perform program transformations, we can recursively traverse this nested list which can be thought of as an abstract syntax tree of the program.

Our program transformations also make use of the Anglican forms \lsi{store} and \lsi{retrieve}.  These allow storing any variable in the probabilistic program's execution trace in a state which is passed around during execution and from which we can retrieve these stored values.  The core use for this is to allow the outer query to return variables which are only locally scoped.

To allow for the early termination that will be introduced in Section \ref{sec:bopp-supp/early-term}, it was necessary to add a mechanism for non-local returns to Anglican.  Clojure supports non-local returns only through Java exception handling, via the keywords {\bf\ttfamily\color{cyan} try}~{\bf\ttfamily\color{cyan}throw},~{\bf\ttfamily\color{cyan}catch} and {\bf\ttfamily\color{cyan}finally}.  Unfortunately, these are not currently supported by Anglican and their behaviour is far from ideal for our purposes.  In particular, for programs containing nested {\bf\ttfamily\color{cyan}try} statements, throwing to a particular {\bf\ttfamily\color{cyan}try} in the stack, as opposed to the most recently invoked, is cumbersome and error prone.
%

We have instead, therefore, added to Anglican a non-local return mechanism based on the Common Lisp control form \lsi{catch/throw}.  This uses a \emph{catch tag} to link each \lsi{throw} to a particular \lsi{catch}.  For example
\begin{lstlisting}[basicstyle=\footnotesize\ttfamily]
(catch :tag
  (when (> a 0)
    (throw :tag a))
  0)
\end{lstlisting}
is equivalent to \lsi{(max a 0)}.  More precisely, \lsi{throw} has syntax \lsi{(throw tag value)} and will cause the \lsi{catch} block with the corresponding \lsi{tag} to exit, returning \lsi{value}.   If a \lsi{throw} goes uncaught, i.e. it is not contained within a \lsi{catch} block with a matching tag, a custom Clojure exception is thrown.


\subsection{Representations in the Main Paper}
\label{sec:bopp-supp/main-paper-rep}

In the main paper we presented the code transformations as static transformations as shown in Figure~\ref{fig:bopp_overview}.  Although for simple programs, such as the given example, these transformations can be easily expressed as static transformations, for more complicated programs it would be difficult to actually implement these as purely static generic transformations in a higher-order language.  Therefore, even though all the transformations dynamically execute as shown at runtime, in truth, the generated source code for the prior and acquisition transformations varies from what is shown and has been presented this way in the interest of exposition.  Our true transformations exploit \lsi{store}, \lsi{retrieve}, \lsi{catch} and \lsi{throw} to generate programs that dynamically execute in the same way at run time as the static examples shown, but whose actual source code varies significantly.

\subsection{Prior Transformation}
\label{sec:bopp-supp/prior-transformations}
The prior transformation recursively traverses the program tree and applies two local transformations.  
Firstly it replaces all \observe statements by \lsi{nil}.  
As \observe statements return \lsi{nil}, this trivially preserves the generative model of the program, but the probability of the execution changes. 
Secondly, it inspects the binding variables of \lsi{let} forms in order to modify the binding expressions for the optimization variables, as specified by the second input of \defopt, asserting that these are directly bound to a \sample statement of the form \texttt{(\sample dist)}.
The transformation then replaces this expression by one that stores the result of this sample in Anglican's \lsi{store} before returning it.
Specifically, if the binding variable in question is \lsi{phi-k}, then the original binding expression \lsi{(sample dist)} is transformed into
    \begin{lstlisting}[basicstyle=\footnotesize\ttfamily]
(let [value (sample dist)]
  ;; Store the sampled value in Anglican's store
  (store OPTIM-ARGS-KEY
         'phi-k
         value)
  value)
    \end{lstlisting}

After all these local transformation have been made, we wrap the resulting query block in a \lsi{do} form and append an expression extracting the optimization variables using Anglican's \lsi{retrieve}.  This makes the optimization variables the output of the query.  Denoting the list of optimization variable symbols from \defopt as \lsi{optim-args} and the query body after applying all the above location transformations as \dots, the prior query becomes
    \begin{lstlisting}[basicstyle=\footnotesize\ttfamily]
(query query-args
  (do
    ...
    (map (fn [x] (retrieve OPTIM-ARGS-KEY x))
       optim-args)))
    \end{lstlisting}
Note that the difference in syntax from Figure~\ref{fig:bopp_overview} is because \lsi{defquery} is in truth a syntactic sugar allowing users to bind \lsi{query} to a variable.  As previously stated, \lsi{query} is macro that compiles an Anglican program to its CPS transformation.  An important subtlety here is that the order of the returned samples is dictated by \lsi{optim-args} and is thus independent of the order in which the variables were actually sampled, ensuring consistent inputs for the BO package.

We additionally add a check (not shown) to ensure that all the optimization variables have been added to the store, and thus sampled during the execution, before returning.  This ensures that our assumption that each optimization variable is assigned for each execution trace is satisfied.

\subsection{Acquisition Transformation}
\label{sec:bopp-supp/acq-transformations}
The acquisition transformation is the same as the prior transformation except we append the acquisition function, \lsi{ACQ-F}, to the inputs and then \observe its application to the optimization variables before returning.
The acquisition query is thus
    \begin{lstlisting}[basicstyle=\footnotesize\ttfamily]
(query [query-args ACQ-F]
  (do
    ...
    (let [theta (map (fn [x] (retrieve OPTIM-ARGS-KEY x))
                      optim-args)]
      (observe (factor) (ACQ-F theta))
      theta)))
    \end{lstlisting}

\subsection{Early Termination}
\label{sec:bopp-supp/early-term}
To ensure that \lsi{q-prior} and \lsi{q-acq} are cheap to evaluate and that the latter does not include unnecessary terms which complicate the optimization, we wish to avoid executing code that is not required for generating the optimization variables.
Ideally we would like to directly remove all such redundant code during the transformations.
However, doing so in a generic way applicable to all possible programs in a higher order language represents a significant challenge.
Therefore, we instead transform to programs with additional early termination statements, triggered when all the optimization variables have been sampled.  
Provided one is careful to define the optimization variables as early as possible in the program (in most applications, e.g. hyperparameter optimization, they naturally occur at the start of the program), this is typically sufficient to ensure that the minimum possible code is run in practise.

To carry out this early termination, we first wrap the query in a \lsi{catch} block with a uniquely generated tag.  We then augment the transformation of an optimization variable's binding described in Section~\ref{sec:bopp-supp/prior-transformations} to check if all optimization variables are already stored, and invoke a \lsi{throw} statement with the corresponding tag if so.  Specifically we replace relevant binding expressions \lsi{(sample dist)} with
    \begin{lstlisting}[basicstyle=\footnotesize\ttfamily]
(let [value (sample dist)]
  ;; Store the sampled value in Anglican's store
  (store OPTIM-ARGS-KEY
         'phi-k
         value)
  ;; Terminate early if all optimization variables are sampled
  (if (= (set (keys (retrieve OPTIM-ARGS-KEY)))
         (set optim-args))
    (throw BOPP-CATCH-TAG prologue-code)
    value))
    \end{lstlisting}
where \lsi{prologue-code} refers to one of the following expressions depending on whether it is used for a prior or an acquisition transformation
    \begin{lstlisting}[basicstyle=\footnotesize\ttfamily]
;; Prior query prologue-code
(map (fn [x] (retrieve OPTIM-ARGS-KEY x))
             optim-args)

;; Acquisition query prologue-code
(do
  (let [theta (map (fn [x] (retrieve OPTIM-ARGS-KEY x))
                    optim-args)]
  (observe (factor) (ACQ-F theta))
  theta))
    \end{lstlisting}

We note that valid programs for both \lsi{q-prior} and \lsi{q-acq} should always terminate via one of these early stopping criteria and therefore never actually reach the appending statements in the \lsi{query} blocks shown in Sections \ref{sec:bopp-supp/prior-transformations} and \ref{sec:bopp-supp/acq-transformations}.  As such, these are, in practise, only for exposition and error catching.

\subsection{Marginal/MMAP Transformation}
The marginal transformation inspects all \lsi{let} binding pairs and if a binding variable \lsi{phi-k} is one of the optimization variables, the binding expression \lsi{(sample dist)} is transformed to the following
    \begin{lstlisting}[basicstyle=\footnotesize\ttfamily]
(do (observe dist phi-k-hat)
    phi-k-hat)
    \end{lstlisting}
corresponding to the \lsi{observe<-} form used in the main paper.

\subsection{Error Handling}
During program transformation stage, we provide three error-handling mechanisms to enforce the restrictions on the probabilistic programs described in Section~\ref{sec:problem}.
\begin{enumerate}
    \item We inspect \lsi{let} binding pairs and throw an error if an optimization variable is bound to anything other than a \sample statement.
    \item We add code that throws a runtime error if any optimization variable is assigned more than once or not at all.
    \item We recursively traverse the code and throw a compilation error if \sample statements of different base measures are assigned to any optimization variable.  At present, we also throw an error if the base measure assigned to an optimization variable is unknown, e.g. because the distribution object is from a user defined \lsi{defdist} where the user does not provide the required measure type meta-information.
\end{enumerate}

%% file: gp.tex

Remembering that the domain scaling introduced in Section~\ref{sec:domain} means that both the input and outputs of the GP are taken to vary between $\pm1$, we define the problem independent GP hyperprior as $p(\alpha)=p(\sigma_n)p(\sigma_{3/2})p(\sigma_{5/2})\prod_{i=1}^{D}p(\rho_i)p(\varrho_i)$ where
\begin{subequations}
	\begin{align}
	\label{eq:hyperPriorDef}
	\log \left(\sigma_n\right) & \sim \mathcal{N} \left(-5,2\right) \\
	\log\left(\sigma_{3/2}\right) & \sim \mathcal{N} \left(-7,0.5\right)\\
	\log\left(\sigma_{5/2}\right) & \sim \mathcal{N} \left(-0.5,0.15\right)\\
	\log \left(\rho_i\right) & \sim \mathcal{N} (-1.5,0.5) \quad \forall i \in \{1,\dots,D\}\\
	\log\left(\varrho_i\right) & \sim \mathcal{N} \left(-1,0.5\right) \quad \forall i \in \{1,\dots,D\}.
	\end{align}
\end{subequations}
The rationale of this hyperprior is that the smoother Mat\'{e}rn 5/2 kernel should be the dominant effect and model the higher length scale variations. The Mat\'{e}rn 3/2 kernel is included in case the evidence suggests that the target is less smooth than can be modelled with the Mat\'{e}rn 5/2 kernel and to provide modelling of smaller scale variations around the optimum.

%% file: house-heating.tex

In this case study, illustrated in Figure~\ref{fig:houses}, we optimize the parameters of a stochastic engineering simulation. We use the Energy2D system from \cite{xie2012energy2d} to perform finite-difference numerical simulation of the heat equation and Navier-Stokes equations in a user-defined geometry. 

In our setup, we designed a 2-dimensional representation of a house with 4 interconnected rooms using the GUI provided by Energy2D. The left side of the house receives morning sun, modelled at a constant incident angle of $30^\circ$. We assume a randomly distributed solar intensity and simulate the heating of a cold house in the morning by 4 radiators, one in each of the rooms. The radiators are given a fixed budget of total power density $P_{\text{budget}}$. The optimization problem is to distribute this power budget across radiators in a manner that minimizes the variance in temperatures across 8 locations in the house. 

Energy2D is written in Java, which allows the simulation to be integrated directly into an Anglican program that defines a prior on model parameters and an ABC likelihood for evaluating the utility of the simulation outputs. Figure \ref{fig:house-heating-code} shows the corresponding program query. In this, we define a Clojure function \lsi{simulate} that accepts a solar power intensity $I_{\text{sun}}$ and power densities for the radiators $P_{\text{r}}$, returning the thermometer temperature readings $\{T_{i, t}\}$. We place a symmetric Dirichlet prior on $\frac{P_r}{P_{\text{budget}}}$ and a gamma prior on $\frac{I_{\text{sun}}}{I_{base}}$, where $P_{\text{budget}}$ and $I_{base}$ are constants. This gives the generative model:
 \begin{align}
 p_r &\sim \Dirichlet([1,1,1,1]) \\
 P_r &\leftarrow P_{\text{budget}} \cdot p_r \\
 \upsilon &\sim \text{Gamma}(5,1) \\
 I_{\text{sun}} &\leftarrow I_{\text{base}} \cdot \upsilon.
 \end{align}
After using these to call \lsi{simulate}, the standard deviations of the returned temperatures is calculated for each time point,
\begin{align}
\omega_t = \sqrt{\sum_{i=1}^8 T_{i, t}^2 -\left(\sum_{i=1}^8 T_{i, t}\right)^2}
\end{align}
and used in the ABC likelihood \lsi{abc-likelihood} to weight the execution trace using a multivariate Gaussian:
\begin{align*}
p\left(\{T_{i, t}\}_{i = 1:8, t = 1:\tau}\right) = \text{Normal}\left(\omega_{t=1:\tau};\mathbf{0},\sigma_T^{2}\mathbf{I}\right)
\end{align*}
where $\mathbf{I}$ is the identity matrix and $\sigma_T = 0.8 ^{\circ} \mathrm{C}$ is the observation standard deviation.

Figure \ref{fig:houses} demonstrates the improvement in homogeneity of temperatures as a function of total number of simulation evaluations. Visual inspection of the heat distributions also shown in Figure \ref{fig:houses} confirms this result, which serves as an exemplar of how BOPP can be used to estimate marginally optimal simulation parameters.